\documentclass[10pt]{article}
\usepackage{fullpage}

\usepackage{graphicx}
\usepackage[dvipsnames]{xcolor}
\usepackage{amsmath}
\usepackage{amssymb}
\usepackage{mathtools}

\makeatletter
\let\save@mathaccent\mathaccent
\newcommand{\let@mathaccent}[1]{\def\composite@test{#1}}
\makeatother
\newcommand{\widebar}[1]{\mkern 1.5mu\overline{\mkern-1.5mu#1\mkern-1.5mu}\mkern 1.5mu}

\usepackage{parskip}
\usepackage{dirtytalk}
\usepackage{algorithm}
\usepackage{algpseudocode}
\usepackage{float}
\usepackage{tikz}
\usetikzlibrary{decorations.pathmorphing}
\usetikzlibrary{decorations.pathreplacing}
\usepackage{siunitx}
\usepackage{bm}
\usepackage{caption}
\usepackage[colorlinks=true,linkcolor=blue]{hyperref}
\usepackage{booktabs}
\usepackage{makecell}
\usepackage{multirow}
\usepackage{longtable}
\usepackage{cleveref}
\usepackage{placeins}

\usepackage[style=authoryear, backend=biber]{biblatex}
\newcommand{\citep}[1]{\parencite{#1}}
\newcommand{\citet}[1]{\textcite{#1}}

\definecolor{malgared}{rgb}{0.8627450980392157, 0.34901960784313724, 0.34901960784313724}

\newtheorem{theorem}{Theorem}[section]

\newtheorem{remark}[theorem]{Remark}

\newcommand{\falkon}{\textsc{Falkon}}
\newcommand{\rpchol}{\textsc{RPCholesky}}
\newcommand{\krill}{\textsc{Krill}}
\newcommand{\eigenpro}[1]{\textsc{EigenPro~#1}}
\newcommand{\joker}{\textsc{Joker}}
\newcommand{\skotch}{\textsc{Skotch}}
\newcommand{\askotch}{\textsc{ASkotch}}
\newcommand{\sketchysaga}{\textsc{SketchySAGA}}
\newcommand{\precfuncname}{\texttt{prec}}
\newcommand{\algname}{\textsc{Bloks}}
\newcommand{\bj}{\textsc{Block~Jacobi}}
\newcommand{\bjf}{\bj\textsc~{-}~\falkon}

\newcommand{\bbR}{\mathbb{R}}
\newcommand{\bigO}{\mathcal{O}}

\newcommand{\kernel}{k}
\newcommand{\cv}{a}                 % coefficient vector
\newcommand{\cvn}{\widetilde{\cv}}  % coefficient vector Nystr\"om
\newcommand{\cvne}{\cvn_\ast}       % coefficient vector Nystr\"om exact solution
\newcommand{\cvnp}{\widetilde{b}}   % coefficient vector Nystr\"om preconditioned problem
\newcommand{\stepsize}{\gamma}

\newcommand{\Knn}{\widehat{K}}
\newcommand{\Knm}{K_{nm}}
\newcommand{\Kmm}{\widetilde{K}}
\newcommand{\Knmj}{K_{nm_0}^{(j)}}
\newcommand{\KnmjT}{K_{nm_0}^{(j)\top}}
\newcommand{\Kmmj}{K_{mm_0}^{(j)}}
\newcommand{\KmmjT}{K_{mm_0}^{(j)\top}}
\newcommand{\Kmjmj}{\Kmm_j}

\newcommand{\Kmzmz}{\Kmm_0}
\newcommand{\Kcomplement}{K_{(n-m_0), m_0}}

\newcommand{\nysmat}{H}
\newcommand{\nysmatnys}{\widetilde{\nysmat}}
\newcommand{\precond}{M}
\newcommand{\preccf}{B}     % Cholesky factor of the preconditioner
\newcommand{\precmatlr}{\preccf^\top \nysmat \preccf}

\DeclareMathOperator{\cond}{cond}
\DeclareMathOperator{\chol}{chol}

\DeclareMathOperator{\relerr}{relerr}
\newcommand{\relerrmat}{\relerr_\nysmat}
\DeclareMathOperator{\relres}{relres}
\newcommand{\relresprec}{\relres_\precond}

\newcommand{\loss}{\widehat{L}}
\newcommand{\lossnys}{\widetilde{L}}

\newcommand{\damping}{\mu}

\newcommand{\na}{N/A}

\title{BLOKS: Scaling Kernel Solvers with Sketched Block Preconditioning}
\author{%
    Giordano Vitale\textsuperscript{1}, Giacomo Meanti\textsuperscript{1}, Marco Rando\textsuperscript{2},\\Blanca Ayuso de Dios\textsuperscript{3,4}, Lorenzo Rosasco\textsuperscript{1,5}\\[1em]
    \small\textsuperscript{1}MaLGa Center, DIBRIS, Università degli Studi di Genova, Genoa, Italy\\
    \small\textsuperscript{2}Université Côte d’Azur, INRIA, CNRS, LJAD, Nice, France\\
    \small\textsuperscript{3}Dipartimento di Matematica e Applicazioni, Università degli Studi Milano–Bicocca, Italy\\
    \small\textsuperscript{4}Istituto di Matematica Applicata e Tecnologie Informatiche \say{E. Magenes}, CNR, Pavia, Italy\\
    \small\textsuperscript{5}SAIL Unit, Istituto Italiano di Tecnologia, Genoa, Italy\\[0.5em]
    \small\texttt{giordano.vitale@edu.unige.it}, \texttt{giacomo.meanti@edu.unige.it}, \texttt{marco.rando@univ-cotedazur.fr},\\
    \small\texttt{blanca.ayuso@unimib.it}, \texttt{lorenzo.rosasco@unige.it}
}
\date{}

\begin{document}

\maketitle

\begin{abstract}

Kernel Ridge Regression (KRR) is a theoretically sound algorithm for nonparametric learning, which is also often used to investigate properties of deep learning. 
However, the exact KRR formulation incurs time and memory costs  preventing applicability to large datasets, motivating the development of approximate solvers. 
These algorithms rely on (i) iterative methods, to avoid the computational burden of direct solutions; (ii) sketching or batching, to reduce the problem size; and (iii) preconditioning, to accelerate convergence. 
While crucial for fast convergence, preconditioning incurs memory costs that can be managed by lowering the number of free parameters of the model, potentially at the expense of expressivity and hence accuracy.
To address this challenge, we propose \algname{}, a Conjugate Gradient-based solver for Nystr\"om-sketched KRR built around a novel preconditioner that combines a block Jacobi approximation with sketching. 
The \algname{} preconditioner has memory cost independent of both the dataset size and the number of parameters, allowing scaling to large models.
We provide a memory-efficient, GPU-accelerated implementation and show that \algname{} achieves excellent accuracy while retaining favorable time/memory complexity. Indeed, on standard large-scale datasets used to benchmark kernel solvers, \algname{} achieves comparable or better performances than state-of-the-art methods.

\end{abstract}

\section{Introduction}
Kernel methods are a well-established class of machine learning models, and their strong theoretical foundations~\citep{steinwartbook} make them a valuable tool for understanding the behavior of deep neural networks~\citep{belkin_understand_dl,belkin_reconc,Jacot_ntk,zhang_under_dl}.  
Despite these advantages, scaling kernel methods to large datasets remains challenging. Training  Kernel Ridge Regression (KRR)  requires solving an $n \times n$ linear system, where $n$ is the number of training points, and computing its exact solution costs $\bigO(n^3)$ time and $\bigO(n^2)$ memory~\citep{scholkopf_smola}, which quickly becomes prohibitive as $n$ grows.

To overcome these limitations, several algorithms have been proposed. They typically compute an approximate solution by combining three main ingredients: iterative methods, sketching or batching, and preconditioning. Existing methods differ in how they combine these ingredients and can be broadly divided into two classes, depending on whether they solve the full problem or a sketched approximation of it. \\
The first class of methods solves the full KRR problem and reduces the time cost through the choice of the optimizer or the design of the preconditioner. Some combine the Conjugate Gradient (CG) method with a sketched preconditioner, retaining the favorable convergence of CG while further reducing the number of iterations \citep{nys_pcg,rff_pcg,krill}. However, each iteration still costs $\bigO(n^2)$ time, limiting these methods to datasets of moderate size. Others reduce the per-iteration cost by relying on stochastic solvers \citep{eigenpro1,eigenpro2} or block-coordinate methods \citep{askotch}, at the price of slower convergence. \\
The second class of methods reduces the number of model parameters to $m \ll n$, using sketching techniques, like  the Nystr\"om method \citep{nystrom,less_is_more} or random Fourier features \citep{rff}, making  each CG iteration much cheaper.  Then,  carefully designed preconditioners for the sketched problem ensure fast convergence \citep{falkon_paper1,falkon_paper2,krill}. However, these methods must store an $m \times m$ preconditioner, whose $\bigO(m^2)$ memory cost becomes prohibitive when $m$ is large \citep{eigenpro3}. This cost can be managed in two ways. The first is to keep $m$ small, which is a significant limitation, since large models are often needed to reach optimal accuracy \citep{eigenpro4,joker,askotch}. The second is to use stochastic solvers with cheaper preconditioners \citep{eigenpro3,eigenpro4} or block-coordinate approaches \citep{joker}, giving up the fast convergence of CG.

In this work, we show that this trade-off can be overcome. We propose \algname{}, a CG-based solver for Nystr\"om-sketched KRR built around a novel preconditioner that combines a block Jacobi approximation with sketching. The \algname{} preconditioner is inexpensive in memory to compute, and its spatial complexity is independent of both the number of data $n$ and the model size $m$. This allows \algname{} to scale to large models while retaining the fast and step-size-free convergence of preconditioned CG. We provide a memory-efficient, GPU-accelerated implementation, allowing us to scale Nystr\"om KRR up to $m = 3.2\mathrm{M}$  parameters. Our numerical experiments show that \algname{} achieves excellent  accuracy while retaining favorable time/memory complexity. 
Indeed, on  standard large-scale datasets used to benchmark kernel solvers, \algname{} achieves comparable or better performances than state-of-the-art methods.

The paper is organized as follows. Section~\ref{sec:background} introduces the setting and the necessary background. Section~\ref{sec:algorithm} introduces the \algname{} algorithm and Section~\ref{sec:related_works} discusses related work. Section~\ref{sec:experiments} reports our experiments, and Section~\ref{sec:conclusion} contains some final remarks.

\section{Background and setting}
\label{sec:background}
Let $\{(x_i, y_i)\}_{i=1}^n \subseteq \bbR^d \times \bbR$ be a finite set of labeled data. Let $\mathcal{H}$ be a reproducing kernel Hilbert space (RKHS) of functions $\bbR^d \to \bbR$ with kernel $\kernel$~\citep{scholkopf_smola}. In KRR, we learn a function $f \in \mathcal{H}$ by solving
\begin{equation}
\label{eq:KRR_functional}
    \min_{f \in \mathcal{H}} \loss(f), \quad \loss(f) = \frac{1}{n} \sum_{i=1}^n (y_i - f(x_i))^2 + \lambda \| f \|_\mathcal{H}^2,
\end{equation}
where $\lambda > 0$ is a regularization parameter and the term $\lambda \| f \|_\mathcal{H}^2$ is added to prevent overfitting. In principle problem~\eqref{eq:KRR_functional} requires minimizing a functional over the possibly infinite dimensional RKHS $\mathcal{H}$. The representer theorem~\citep{representer1, representer2} however states that the minimum is attained by functions of the form
\begin{equation}
    f_\cv(x) = \sum_{i=1}^n \cv_i \kernel(x, x_i),
\end{equation}
parametrized by a vector of coefficients $\cv = (\cv_1, \dots, \cv_n)^\top \in \bbR^n$. By substituting the expression of $f_a$ into $\loss(f)$, KRR reduces to the finite dimensional minimization problem
\begin{equation}
\label{eq:KRR}
    \min_{\cv \in \bbR^n} \loss(\cv), \qquad \loss(\cv) = \frac{1}{n} \| \widehat{y} - \Knn \cv \|^2 + \lambda a^\top \widehat{K} a,
\end{equation}
where $\widehat{y} = (y_1, \dots, y_n)^\top \in \bbR^n$, $\Knn \in \bbR^{n \times n}$ is the kernel matrix defined by $(\Knn)_{ij} = \kernel (x_i, x_j)$ for $i, j = 1, \dots, n$ and $\| \cdot \|$ denotes the standard Euclidean norm. Solving problem \eqref{eq:KRR} is equivalent to solving the linear system
\begin{equation}
\label{eq:KRR_system}
    (\Knn + \lambda nI)\cv = \widehat{y}.
\end{equation}
In large scale scenarios, directly solving problem \eqref{eq:KRR_system} is unfeasible. Performing the Cholesky decomposition of $\Knn + \lambda n I$ costs $\bigO(n^3)$ in time and $\bigO(n^2)$ in memory~\citep{golub_van_loan}. Modern algorithms typically address this computational cost through three approaches: {\bf sketching}, {\bf iterative methods} and {\bf preconditioning}.

\paragraph{Sketching.} As the matrix $\Knn + \lambda n I$ is too large to effectively operate with, the first idea is to compress, or \emph{sketch} it into a smaller matrix. Popular techniques for kernel matrix compression include Random Fourier Features (RFF)~\citep{rff} and Nystr\"om-based approximations~\citep{nystrom}. We focus here on the Nystr\"om approach. Let $m \leq n$ and $\{\widetilde{x}_1, \dots, \widetilde{x}_m\} \subseteq \{x_1, \dots, x_n\}$ sampled uniformly at random. We call $\{\widetilde{x}_1, \dots, \widetilde{x}_m\}$ the Nystr\"om centers and consider predictors of the form
\begin{equation}
    \widetilde{f}_{\cvn}(x) = \sum_{i=1}^m \cvn_i \kernel(x, \widetilde{x}_i),
\end{equation}
where $\cvn = (\cvn_1, \dots, \cvn_m)^\top \in \bbR^m$ is a compressed vector of coefficients. By restricting problem \eqref{eq:KRR_functional} to functions of this form, Nystr\"om KRR consists of solving
\begin{equation}
\label{eq:NKRR}
    \min_{\cvn \in \bbR^m} \lossnys(\cvn), \quad \lossnys(\cvn) = \frac{1}{n} \| \widehat{y} - \Knm \cvn \|^2 + \lambda \cvn^\top \Kmm \cvn,
\end{equation}
where $\Knm \in \bbR^{n \times m}$ and $\Kmm \in \bbR^{m \times m}$ are kernel matrices defined respectively by $(\Knm)_{ij} = \kernel (x_i, \widetilde{x}_j)$ and $(\Kmm)_{jj'} = \kernel (\widetilde{x}_j, \widetilde{x}_{j'})$ for $i = 1, \dots, n$ and $j, j' = 1, \dots, m$. Solving \eqref{eq:NKRR} is equivalent to solving the linear system
\begin{equation}
\label{eq:NKRR_system}
    \underbrace{(\Knm^\top \Knm + \lambda n \Kmm)}_{\eqqcolon \nysmat}\cvn = \Knm^\top \widehat{y}.
\end{equation}
\begin{remark}[Invertibility of $\Kmm$]
    For simplicity, we assume the matrix $\Kmm$ to be invertible. In practice this assumption is ensured by means of jittering.\footnote{In general $\Kmm$ is a symmetric and positive semidefinite matrix. We fix a small jittering parameter $\varepsilon > 0$ and perturb $\Kmm + \varepsilon I$ to ensure positive definiteness, and therefore invertibility. The magnitude of $\varepsilon$ is chosen relative to the working numerical precision.} Nonetheless, system \eqref{eq:NKRR_system} is always consistent since $\Knm^\top \widehat{y}$ is always in the range of $\nysmat = \Knm^\top \Knm + \lambda n \Kmm$, meaning a solution always exists.
\end{remark}
Solving~\eqref{eq:NKRR_system} directly is still unfeasible in large scale scenarios, due to the prohibitive cost of $\bigO(nm^2)$ and numerical instability of the Gram product $\Knm^\top \Knm$. Additionally, explicitly storing the matrix $\nysmat$ requires $\bigO(m^2)$ memory, which becomes prohibitive in settings where large values of $m$ are needed \citep{eigenpro4}. %This requirement is overly restrictive, as our objective is scaling to large values of $m$. 
To tackle such limitations, an alternative approach is to rely on iterative methods~\citep{saad}.

\paragraph{Iterative methods.} These are procedures that produce a sequence of approximations $(\cvn_t)_{t\in \mathbb{N}}$ converging  to the exact solution of~\eqref{eq:NKRR_system} without inverting the matrix $\nysmat$ explicitly. To illustrate this idea, consider the Gradient Descent (GD) iteration % for minimizing~\eqref{eq:NKRR}
\begin{equation}
\label{eq:GD}
    \cvn_{t+1} = \cvn_t + \stepsize (\Knm^\top \widehat{y} - \nysmat \cvn_t) \qquad t \geq 0,
\end{equation}
for a fixed initial guess $\cvn_0 \in \bbR^m$ and step size $\stepsize > 0$. Notice that~\eqref{eq:GD} only requires the action of $\nysmat$, which can be computed in arbitrarily small blocks, in principle completely nullifying the memory footprint of the algorithm.\footnote{Note however that while reducing block size mitigates spatial complexity, it introduces a significant temporal overhead, due to the more frequent data swapping between secondary storage and system (RAM) or device (GPU) memory.}
Performing $t_{\max}$ iterations of GD requires $\mathcal{O}(t_{\max}(nm + m^2))$ time, compared to $\mathcal{O}(nm^2 + m^3)$ for explicitly forming and inverting $\nysmat$. On the one hand, this motivates the use of a small number of iterations $t_{\max} \ll m$ to reduce the overall cost; on the other hand, too few iterations may result in a poor approximation of the solution of~\eqref{eq:NKRR_system}. For iterative methods, the number of iterations required to achieve a certain accuracy in approximating the exact solution depends on the convergence rate. Notably, for gradient-based methods (as GD), this rate depends on the condition number~\citep{golub_van_loan} of $\nysmat$, motivating the introduction of a preconditioner.

\paragraph{Preconditioning.} We begin by recalling the definition of the condition number of $\nysmat$
\begin{equation}
\cond(\nysmat) = \frac{\lambda_{\max}(\nysmat)}{\lambda_{\min}(\nysmat)},
\end{equation}
where $\lambda_{\max}(\nysmat)$ and $\lambda_{\min}(\nysmat)$ denote the largest and smallest eigenvalues of $\nysmat$, respectively.
Kernel matrices can be highly ill-conditioned, and preconditioning addresses this issue by transforming the original system \eqref{eq:NKRR_system} into an equivalent one with improved condition number. Specifically, given an invertible preconditioner $\precond = \preccf \preccf^\top \in \bbR^{m \times m}$, we consider the system
\begin{equation}
\label{eq:NKRR_system_prec}
    \precmatlr \cvnp = \preccf^\top \Knm^\top \widehat{y}, \qquad \cvn = \preccf \cvnp.
\end{equation}
Designing $\precond$ involves two main considerations: (i) it should be efficient to apply in terms of both time and memory, and (ii) it should improve the conditioning of the system, that is, $\cond(\preccf^\top \nysmat \preccf) < \cond(\nysmat)$. A common approach is to construct $\precond = \nysmatnys^{-1}$ with $\nysmatnys \in \bbR^{m \times m}$ an approximation of $\nysmat$. This is precisely where a memory bottleneck may arise, as storing and inverting $\nysmatnys$ as a full, unstructured matrix requires $\mathcal{O}(m^2)$ memory.

We have seen that through sketching, iterative methods and preconditioning, time and space costs of KRR can be alleviated. We are left with a memory bottleneck in the preconditioner, scaling quadratically in space with the number of parameters of the model. Overcoming this final bottleneck motivates the introduction of \algname{} in the next section.

\section{Introducing BLOKS}
\label{sec:algorithm}
In this section, we introduce \algname{}, a memory-efficient algorithm which overcomes the preconditioner memory bottleneck and scales Nystr\"om KRR to high parameter counts. The core idea is to build a sketched block diagonal preconditioner over disjoint subsets of Nystr\"om centers. This allows us to retain the benefits of preconditioning while allowing efficient action computation (both in memory and time). The \algname{} algorithm uses such a preconditioner within Conjugate Gradient (CG), which has a more favorable dependence on the condition number in the rate than GD and keeps the same computational cost \citep{saad}. To introduce the \algname{} algorithm, we proceed in three steps: (i) we define our sketched block diagonal preconditioner; (ii) we show how its action on vectors can be computed efficiently both in memory and time; (iii) we discuss why the left variant of the Preconditioned Conjugate Gradient (PCG) algorithm is more time efficient in large-scale settings.

\paragraph*{Sketched block Jacobi preconditioning.} Let $1 \leq m_0 \leq m$ such that $m_0$ divides $m$ and let $q = m/m_0$. We partition the Nystr\"om centers in $q$ disjoint blocks of size $m_0$ %$(m_i)_{i=1}^q$ i.e.
\begin{equation}\label{eq:data_partition}
\begin{aligned}
    \underbrace{\widetilde{x}_1, \dots, \widetilde{x}_{m_0}}_{\text{block 1}},\dots, \underbrace{\widetilde{x}_{m - m_0 + 1}, \dots, \widetilde{x}_{m}}_{\text{block q}}.
\end{aligned}
\end{equation}
For each block $j = 1, \dots, q$, we define
\begin{equation}
    % \nysmatnys_j = \frac{nq}{m_j}\Kmjmj^2 + \lambda n q \Kmjmj,
    \nysmatnys_j = \frac{nq}{m_0}\Kmjmj^2 + \lambda n q \Kmjmj,
\end{equation}
where $\Kmjmj \in \bbR^{m_0 \times m_0}$ is the kernel matrix whose entries are obtained by evaluating the kernel on the points of block $j$. Formally, $(\Kmjmj)_{{l_j}{l'_j}} = \kernel(\widetilde{x}_{l_j}, \widetilde{x}_{l'_j})$ for $l_j, l'_j = (j - 1) m_0 + 1, \dots, j m_0$. 
We define the \algname{} preconditioner as
\begin{equation}
\label{eq:blk_prec}
    \precond = \operatorname{diag}\left( \nysmatnys_1^{-1}, \dots, \nysmatnys_q^{-1} \right).
\end{equation}
One core advantage of this preconditioner is that its block diagonal structure allows its action on vectors to be computed in blocks, namely by solving $q$ decoupled linear systems. Given a vector $r \in \mathbb{R}^m$, we partition it in $q$ blocks of size $m_0$ as $r = (r^{(1)}, \dots, r^{(q)})^\top$, coherently with \eqref{eq:data_partition}. Then, the action of $\precond$ on $r$ is computed as
\begin{equation}
    \precond r =
    \begin{bmatrix}
        \nysmatnys_1^{-1} r^{(1)}\\
        \vdots\\
        \nysmatnys_q^{-1} r^{(q)}
    \end{bmatrix}.
\end{equation}
This procedure could be naturally parallelized, especially on high performance systems with multiple GPUs or whenever multiple blocks can be stored in memory simultaneously. In large-scale settings when the block size is too large to allow simultaneous storage, we can process the blocks sequentially, yielding a memory cost of $\mathcal{O}(m_0^2)$, independent of $m$.
Note that the \algname{} preconditioner can be interpreted as a sketched version of the \bj{} preconditioner \citep{saad} for problem \eqref{eq:NKRR_system}, which consists of directly approximating $H$ by its block diagonal and then inverting it.
However, computing the action of each block of plain \bj{} would require computing a Gram product of the form $K_{nm_0}^\top K_{nm_0}$ (see Figure \ref{fig:first_derivation} for an illustration) which is numerically unstable and has a time cost of $\mathcal{O}(nm_0^2)$ which is computationally too inefficient in large-scale scenarios (see Appendix \ref{apx:block_prec} for a detailed comparison). 
\begin{figure*}[ht]
    \centering
    \colorlet{blockgray}{gray!30}
    \begin{tikzpicture}[scale=1,
        seconddiag/.style={thick, dashed}
    ]
    \def\bs{1.2}
    \def\ms{2.4}
    \def\ns{3.6}

    \pgfmathsetmacro{\yTallMin}{\ms/2 - \ns/2}
    \pgfmathsetmacro{\yTallMax}{\ms/2 + \ns/2}

    \def\xMtwo{3.2}
    \def\xMthree{7.0}
    \def\xMfour{10.2}
    
    \def\yLineTwo{-4.2}
    \def\yLineTwoMid{-3.0} 

    \begin{scope}[shift={(0,0)}]
        \fill[blockgray] (0,0) rectangle (\ms,\ms);
        \fill[malgared] (0,\bs) rectangle (\bs,\ms);
        
        \draw[seconddiag] (\bs,0) -- (\bs,\bs) -- (\ms,\bs);

        \draw[thick] (0,0) rectangle (\ms,\ms);
        \draw[thick] (0,\bs) rectangle (\bs,\ms);

        \node at (\bs/2, \bs + \bs/2) {$\propto \nysmat_1$};
        
        \draw[decorate,decoration={brace,amplitude=5pt},thick] 
            (0,\ms+0.05) -- (\bs,\ms+0.05) node[midway, above=6pt] {$m_0$};
        \draw[decorate,decoration={brace,amplitude=5pt},thick] 
            (-0.05,\bs) -- (-0.05,\ms) node[midway, left=6pt] {$m_0$};
            
        \node[below=8pt] at (\bs, 0) {$\nysmat \propto \Knm^\top \Knm + \Kmm$};
    \end{scope}

    \node at (2.8, \bs) {\Large $=$};

    \begin{scope}[shift={(\xMtwo,0)}]
        \fill[blockgray] (0,0) rectangle (\ns,\ms);
        \fill[malgared] (0,\bs) rectangle (\ns,\ms);
        
        \draw[thick] (0,0) rectangle (\ns,\ms);
        \draw[thick] (0,\bs) rectangle (\ns,\ms);
        
        \node at (\ns/2, \bs + \bs/2) {$K_{nm_0}^{(1)\top}$};
    \end{scope}

    \begin{scope}[shift={(\xMthree,0)}]
        \fill[blockgray] (0,\yTallMin) rectangle (\ms,\yTallMax);
        \fill[malgared] (0,\yTallMin) rectangle (\bs,\yTallMax);
        
        \draw[thick] (0,\yTallMin) rectangle (\ms,\yTallMax);
        \draw[thick] (0,\yTallMin) rectangle (\bs,\yTallMax);
        
        \node at (\bs/2, \ms/2) {$K_{nm_0}^{(1)}$};
    \end{scope}

    \node at (9.8, \bs) {\Large $+$};

    \begin{scope}[shift={(\xMfour,0)}]
        \fill[blockgray] (0,0) rectangle (\ms,\ms);
        \fill[malgared] (0,\bs) rectangle (\bs,\ms);
        
        \draw[seconddiag] (\bs,0) -- (\bs,\bs) -- (\ms,\bs);

        \draw[thick] (0,0) rectangle (\ms,\ms);
        \draw[thick] (0,\bs) rectangle (\bs,\ms);
        
        \node at (\bs/2, \bs + \bs/2) {$\Kmm_1$};
    \end{scope}

    \node at (2.8, \yLineTwoMid) {\Large $\approx$};

    \begin{scope}[shift={(\xMtwo,\yLineTwo)}]
        \fill[blockgray] (0,0) rectangle (\ns,\ms);
        \fill[malgared] (0,\bs) rectangle (\bs,\ms);
        
        \draw[seconddiag] (\bs,\bs) -- (\ns,\bs);

        \draw[thick] (0,0) rectangle (\ns,\ms);
        \draw[thick] (0,\bs) rectangle (\bs,\ms);
        
        \node at (\bs/2, \bs + \bs/2) {$\Kmm_1$};
    \end{scope}

    \begin{scope}[shift={(\xMthree,\yLineTwo)}]
        \fill[blockgray] (0,\yTallMin) rectangle (\ms,\yTallMax);
        \fill[malgared] (0,\yTallMax-\bs) rectangle (\bs,\yTallMax);
        
        \draw[seconddiag] (\bs,\yTallMin) -- (\bs,\yTallMax-\bs);

        \draw[thick] (0,\yTallMin) rectangle (\ms,\yTallMax);
        \draw[thick] (0,\yTallMax-\bs) rectangle (\bs,\yTallMax);
        
        \node at (\bs/2, \yTallMax - \bs/2) {$\Kmm_1$};
    \end{scope}

    \node at (9.8, \yLineTwoMid) {\Large $+$};

    \begin{scope}[shift={(\xMfour,\yLineTwo)}]
        \fill[blockgray] (0,0) rectangle (\ms,\ms);
        \fill[malgared] (0,\bs) rectangle (\bs,\ms);
        
        \draw[seconddiag] (\bs,0) -- (\bs,\bs) -- (\ms,\bs);

        \draw[thick] (0,0) rectangle (\ms,\ms);
        \draw[thick] (0,\bs) rectangle (\bs,\ms);
        
        \node at (\bs/2, \bs + \bs/2) {$\Kmm_1$};
    \end{scope}

    \end{tikzpicture}
    \caption{First block of $\nysmat$ (above), and sketched approximation (below). Constants omitted for visual clarity.}
    \label{fig:first_derivation}
\end{figure*}

\paragraph*{Efficient solution with $\precond$.} Unlike standard \bj{}, the action of our sketched preconditioner can be computed efficiently. In fact, each sketched block $\nysmatnys_j^{-1}$ can be expressed as
\begin{equation}
\label{eq:blk_prec_decomposition}
    \begin{aligned}
        \nysmatnys_j^{-1} = \preccf_j \preccf_j^\top, \qquad B_j = \frac{1}{\sqrt{nq}} T_j^{-1} A_j^{-1}, \qquad T_j = \chol(\Kmjmj), \qquad A_j = \chol \left( \frac{1}{m_0}T_j T_j^\top + \lambda I \right),
    \end{aligned}
\end{equation}
where $\chol(\cdot)$ denotes the upper triangular Cholesky factor of a symmetric and positive definite matrix. In this way, we can solve with each block $\nysmatnys_j^{-1}$ without forming the product $\Kmjmj^2$ explicitly.  

Furthermore, computing the factors $T_j$ and $A_j$ is highly memory-efficient, requiring the storage of only a single $m_0 \times m_0$ matrix. Once the matrix $\Kmjmj$ is computed, an in-place Cholesky decomposition stores the factor $T_j$ in the upper triangular portion. The lower triangular portion is then reused to store the symmetric matrix $\frac{1}{m_0}T_jT^\top_j + \lambda I$, which is finally overwritten by $A_j^\top$ by a second in-place Cholesky decomposition. Additional care is required to handle the diagonals of these matrices - see Appendix \ref{apx:pseudocode_long}.

\paragraph*{Left PCG.} The block structure of $\precond$ makes its application particularly suitable for a left-preconditioned formulation. In contrast to the bilateral formulation in equation~\eqref{eq:NKRR_system_prec}, which applies the factors of $M$ to the left and to the right of $\nysmat$, we can precondition from the left by formulating the problem as
\begin{equation}
\label{eq:NKRR_system_prec_left}
\precond \nysmat \cvn = \precond \Knm^\top \widehat{y}.
\end{equation}
This choice is important in memory-constrained settings: when $m_0$ is large and only one block can be stored, applying $\precond$ from the left requires computing each block only once per iteration, while the bilateral formulation requires a second recomputation - further details and comparison are provided in Appendix~\ref{apx:bilateral_vs_left_prec}. Combining thus our preconditioner with left PCG, we get the \algname{} algorithm, summarized in Algorithm \ref{alg:blk_algorithm_short} and detailed in Algorithm~\ref{alg:blk_algorithm_long}.
\begin{table}[ht!]
\centering
\begin{algorithm}[H]
\caption{Pseudocode for the \algname{} algorithm}
\label{alg:blk_algorithm_short}

\begin{algorithmic}[1]
    \Statex \textbf{Input}: data matrix $\widehat{X} = [x_1 \mid \dots \mid x_n]^\top \in \bbR^{n \times d}$, label vector $\widehat{y} \in \bbR^n$, penalty $\lambda > 0$, model size $m$, block size $m_0 \mid m$, kernel $\kernel$, number of iterations $t_{\max}$
    \Statex \textbf{Output}: Nystr\"om predictor $\cvn$ (approximated coefficient vector)
    
    \vspace{0.2cm}

    \State $c = \texttt{sample}(\{1, \dots, n\}, m)$ \label{line:sampling}
    \State $\widetilde{X} = \widehat{X}[c]$
    \State $\Knm = \kernel(\widehat{X}, \widetilde{X})$
    \State $\Kmm = \Knm[c]$
    \State $\nysmat = \Knm^\top \Knm + \lambda n \Kmm$
    \State $\cvn = \texttt{left\_pcg}(\nysmat, K_{nm}^\top\widehat{y}, \precfuncname, t_{\max})$ \Comment{\small Algo.~\ref{alg:LPCG}}
    \State \Return $\cvn$
    
    \vspace{0.5cm}
    
    \Function{\precfuncname}{$r; \Kmm, m_0$} \Comment{\small Computes $\precond r$ (Algo.~\ref{alg:blk_l})}
        \State $q = m / m_0$
            \For{$j = 1$ to $q$}
                \State $T_j = \chol(\Kmjmj)$
                \State $A_j = \chol \left( \frac{1}{m_0}T_j T_j^\top + \lambda I \right)$
                \State $r^{(j)} \gets T_j^{-1}(A_j^{-1}(A_j^{-T}(T_j^{-T} r^{(j)})))$
                \State \textbf{delete} $T_j, A_j$
            \EndFor
        \State \Return $r \gets \frac{1}{nq} r$
    \EndFunction
\end{algorithmic}
\end{algorithm}
\end{table}

\section{Related works}
\label{sec:related_works}
The design of efficient kernel algorithms has been studied extensively in the literature. One line of work improves scalability by splitting the problem into smaller ones, partitioning the data~\citep{divide_and_conquer}, the input space~\citep{input_space_partition}, or the feature space~\citep{carratino2021park}. These strategies are complementary to ours, and here we focus on methods that directly solve the KRR problem, which fall into two main categories based on the problem they address:  the full problem ({\it full KRR methods}) or a sketched approximation ({\it sketching-based methods}). We briefly review the existing methodologies closer to our proposed approach within these categories. Further details and comparisons are provided in Appendix~\ref{apx:further_comparison}.

\paragraph*{Full KRR methods.}
They solve the $n$-dimensional system~\eqref{eq:KRR_system}, either using PCG or stochastic gradient descent (SGD). Methods using PCG  differ mostly on the design of the randomized preconditioner: in \citet{nys_pcg} the preconditioner is built from a Nystr\"om approximation of $\Knn$; in \citet{rff_pcg} from a RFF sketch of $\Knn$ and in \citet{krill} from a randomly pivoted Cholesky approximation~\citep{rpchol}. In all cases, every PCG iteration costs $\bigO(n^2)$ time, limiting the scaling to modest size datasets.
To alleviate the iteration cost, \eigenpro{1} and \eigenpro{2}~\citep{eigenpro1,eigenpro2} use SGD preconditioned using the top eigenvectors of $\Knn$ computed via randomized SVD. The sketch-and-project approach~\citep{sap}  can be regarded as SGD with a random low-rank preconditioner built from a sketch of the system.  In block-coordinate descent the \say{preconditioner} selects a block of coordinates and rescales it by the inverse of the corresponding diagonal block. \textsc{RidgeSketch}~\citep{sketch_ridge} uses count-sketch, and  proposes a variant with momentum. Block coordinate descent is used in \citet{tu2016largescalekernellearning}; in the exact version of \joker{} \citep{joker} (it is used in the dual of problem \eqref{eq:KRR}) and in  \skotch{} and \askotch{}~\citep{askotch}. Momentum acceleration is also  incorporated in \citet{pmlr-v70-tu17a} and \askotch{}.
 
\paragraph*{Sketching-based methods.} These methods design solvers and preconditioners for sketched approximations such as~\eqref{eq:NKRR_system}. Typically the  sketching is done  through RFF \citep{joker} or Nystr\"om approximation.
\eigenpro{3} and \eigenpro{4}~\citep{eigenpro3,eigenpro4} use projected preconditioned SGD on problem~\eqref{eq:NKRR_system} using a randomized SVD-based preconditioner.  \eigenpro{4} reduces the cost of \eigenpro{3} by performing the projection step only periodically. 
\sketchysaga{}~\citep{sketchysaga,askotch} combines SAGA variance reduction technique \citep{saga_defazio,pmlr-v97-gazagnadou19a} with subsampling and randomized Nystr\"om preconditioner.
This mitigates the slow rate of SGD methods while avoiding the cost of full-gradient computation and reduces ill-conditioning through the preconditioner. Although they reduce the memory requirement, they rely either on SGD or on variance reduction techniques which have a slower convergence rate than CG. \falkon{}~\citep{falkon_paper1,falkon_paper2} solves problem~\eqref{eq:NKRR_system} with PCG, using a sketched preconditioner built from the matrix $\Kmm$. \krill{}~\citep{krill} follows the same idea but uses a more refined sketching technique to design the preconditioner. However, both algorithms require $\bigO(m^2)$ memory to store the preconditioner, which becomes prohibitive for a large model size $m$, as observed by \citet{eigenpro3}. In contrast, the \algname{} preconditioner has a fixed memory cost of $\bigO(m_0^2)$, independent of $m$ and $n$, while keeping CG as the backbone solver, which converges faster than GD and stochastic variants, and requires no step size tuning. Notably, we observe that \algname{} recovers \falkon{} as a special case when $m_0 = m$, that is, when there is only $q = 1$ block. Scalability with sketching has also been studied for GPs \citep{quinonero2005unifying}, where a large model size $m$ has proven crucial as well \citep{kiss-gp,burt19,hensman2015scalable,matthews2016sparse}.

\section{Experiments}
\label{sec:experiments}

We compared \algname{} with \eigenpro{4}~\citep{eigenpro4}, \joker{}~\citep{joker}, \askotch{}~\citep{askotch} and \falkon{}~\citep{falkon_paper2} on both large-scale and medium-scale datasets, to analyze their predictive performance, their scalability and their efficiency. In this section we present the overall results and key takeaways; all relevant experimental details are available in Appendix~\ref{apx:experiment_details}, while additional experiments comparing different block preconditioning strategies are in Appendix~\ref{apx:block_prec}.

\begin{table}[t]
\centering
\caption{Large-scale dataset comparisons. For each method/dataset combination we report the training time and a relevant measure of error or accuracy on that dataset. Standard deviations (over 5 data splits) are reported in Appendix~\ref{apx:experiment_details} for space reasons. Bold green cells indicate the best training time and the best accuracy on a dataset; light-green cells indicate accuracy within 1 standard deviation of the best.}
\label{tbl:large-scale}
\small
\setlength{\tabcolsep}{4pt}
\begin{tabular*}{\textwidth}{@{\extracolsep{\fill}}l*{10}{c}@{}}
\toprule
& \multicolumn{2}{c}{\makecell{Taxi\\[-1pt]{\footnotesize($n\approx10^9$)}}} & \multicolumn{2}{c}{\makecell{Higgs\\[-1pt]{\footnotesize($n\approx10^7$)}}} & \multicolumn{2}{c}{\makecell{Airlines\\[-1pt]{\footnotesize($n\approx10^6$)}}} & \multicolumn{2}{c}{\makecell{CIFAR10-5M\\[-1pt]{\footnotesize($n\approx10^6$)}}} & \multicolumn{2}{c}{\makecell{COMET MC\\[-1pt]{\footnotesize($n\approx10^7$)}}} \\
\cmidrule{2-3} \cmidrule{4-5} \cmidrule{6-7} \cmidrule{8-9} \cmidrule{10-11}
Algorithm & time & RMSE & time & AUC & time & MSE & time & accuracy & time & AUC \\
\midrule
\falkon & {\boldmath \color{ForestGreen} $\SI{2}{\hour}$} & ${307.9}$ & {\boldmath \color{ForestGreen} $\SI{7}{\minute}$} & ${0.8194}$ & {\boldmath \color{ForestGreen} $\SI{2}{\minute}$} & ${0.720}$ & {\boldmath \color{ForestGreen} $\SI{14}{\minute}$} & ${87.7\%}$ & {\boldmath \color{ForestGreen} $\SI{46}{\second}$} & ${0.9911}$ \\
\joker & $\SI{7}{\hour}$ & ${307.8}$ & $\SI{47}{\minute}$ & ${0.8200}$ & $\SI{1}{\hour}$ & ${0.718}$ & $\SI{1}{\hour}$ & ${87.7\%}$ & $\SI{56}{\minute}$ & ${0.9923}$ \\
\askotch & -- & -- & $\SI{4}{\hour}$ & ${0.8237}$ & $\SI{30}{\minute}$ & {\color{YellowGreen} ${0.717}$} & -- & -- & $\SI{34}{\minute}$ & {\color{YellowGreen} ${0.9941}$} \\
\eigenpro{4} & -- & -- & $\SI{4}{\hour}$ & ${0.8104}$ & $\SI{2}{\hour}$ & ${0.791}$ & $\SI{1}{\hour}$ & ${86.9\%}$ & $\SI{2}{\hour}$ & ${0.9759}$ \\
\algname & $\SI{5}{\hour}$ & {\boldmath \color{ForestGreen} ${307.0}$} & $\SI{2}{\hour}$ & {\boldmath \color{ForestGreen} ${0.8254}$} & $\SI{47}{\minute}$ & {\boldmath \color{ForestGreen} ${0.706}$} & $\SI{2}{\hour}$ & {\boldmath \color{ForestGreen} ${88.0\%}$} & $\SI{31}{\minute}$ & {\boldmath \color{ForestGreen} ${0.9943}$} \\
\bottomrule
\end{tabular*}
\end{table}

\begin{figure}[t]
\centering
\begin{minipage}[c]{0.49\textwidth}
    \centering
    \includegraphics[width=\linewidth]{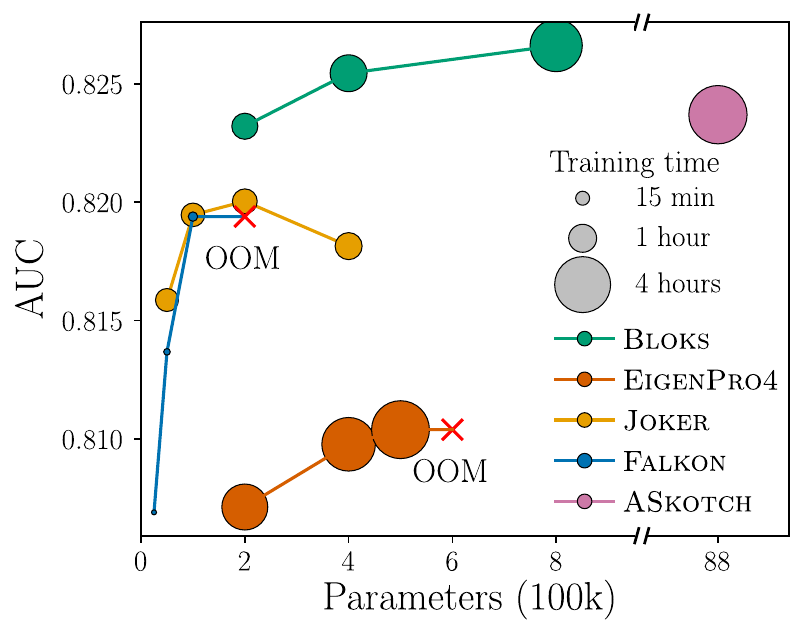}
\end{minipage}%
\hfill
\begin{minipage}[c]{0.45\textwidth}
    \centering
    \renewcommand{\arraystretch}{1.15}
    \begin{tabular}{ccc}
    \toprule
    $m$ & hours & AUC \\
    \midrule
    $200\mathrm{k}$ & $0.87$  & $0.8232$ \\
    $400\mathrm{k}$ & $1.74$  & $0.8254$ \\
    $800\mathrm{k}$ & $3.49$  & $0.8266$ \\
    $1.6\mathrm{M}$ & $7.21$  & $0.8272$ \\
    $3.2\mathrm{M}$ & $14.69$ & $0.8277$ \\
    \bottomrule
    \end{tabular}
\end{minipage}
\par\medskip
\begin{minipage}[t]{0.49\textwidth}
    \caption{AUC as a function of the number of parameters. We tested different algorithms with increasing model sizes on the Higgs dataset, plotting the test performance and the training time (proportional to marker area). We mark the algorithms which run out of memory (OOM).}
    \label{fig:higgs-scale}
\end{minipage}%
\hfill
\begin{minipage}[t]{0.45\textwidth}
    \captionof{table}{\algname{} scaling of the model size $m$ on the HIGGS dataset. Training time scales linearly, and the test AUC improves consistently. Standard deviations (over 5 data splits) are reported in Appendix~\ref{apx:experiment_details}.}
    \label{tab:higgs-scale}
\end{minipage}
\end{figure}

\begin{figure}[t]
\centering
\begin{minipage}[t]{0.48\textwidth}
    \centering
    \includegraphics[width=\linewidth]{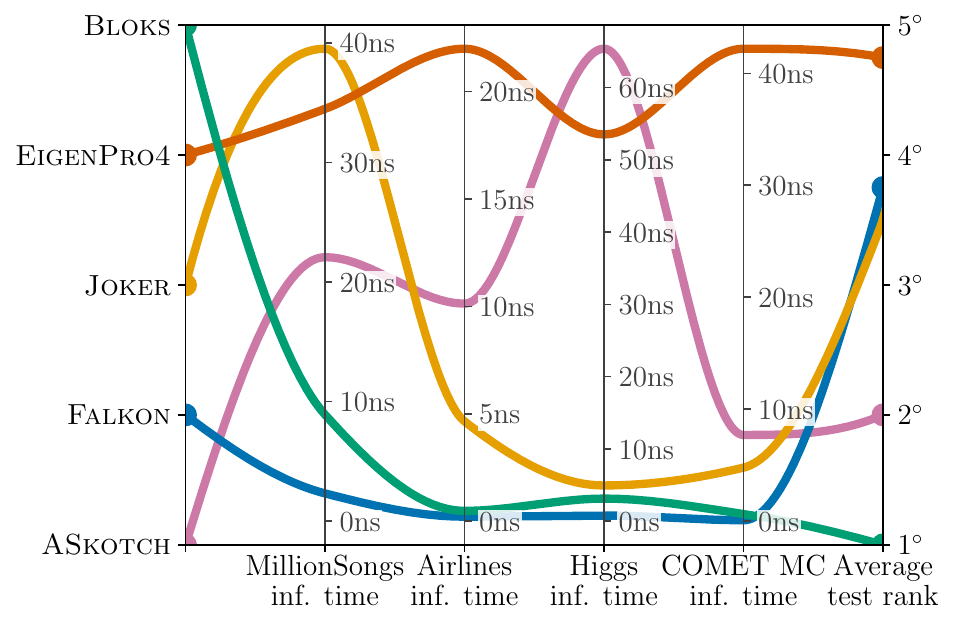}
    \caption{Inference times and average test performance. We compare the inference time (normalized per test-sample) across algorithms and datasets. The rightmost column shows the average test ranking of the algorithms across the four datasets considered.}
    \label{fig:inf-time}
\end{minipage}%
\hfill
\begin{minipage}[t]{0.48\textwidth}
    \centering
    \includegraphics[width=\linewidth]{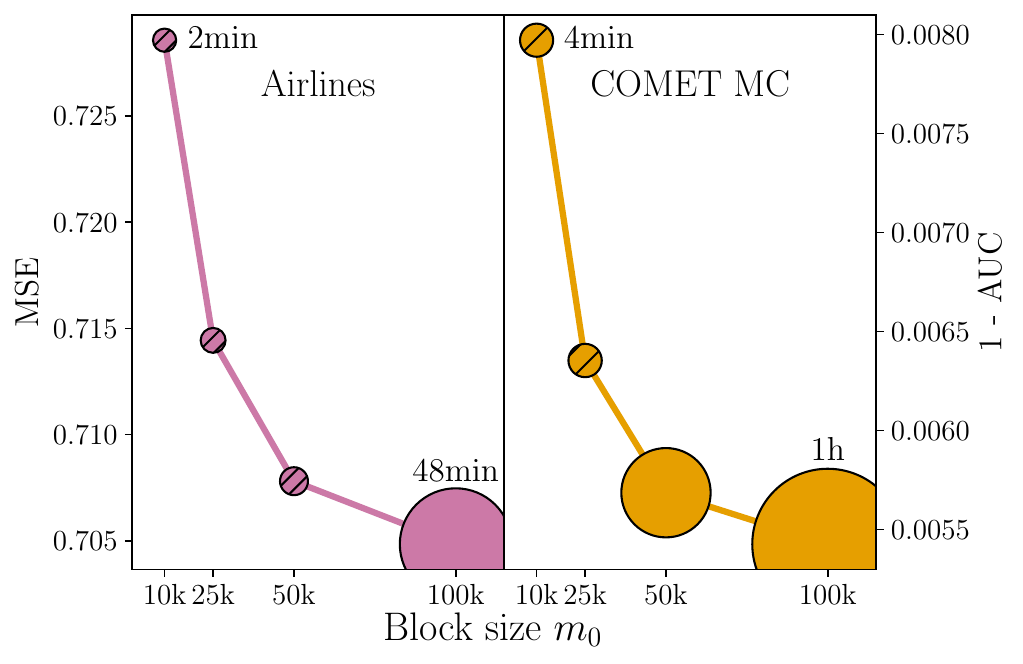}
    \caption{Test errors and training times with increasing block sizes. For both \emph{Airlines} and \emph{COMET~MC} datasets increasing $m_0$ decreases the test error while increasing the training time (proportional to the marker area). Hatched markers indicate runs in which all blocks could be stored in memory.}
    \label{fig:ablation-inc-m0}
\end{minipage}
\end{figure}

\paragraph{Parameter scaling.}
We begin by evaluating the algorithms' scalability with the number of model parameters. This relates to one of the core aspects of \algname{} which decouples model size and memory requirements. As we show in \Cref{fig:higgs-scale} on the \emph{Higgs} dataset ($n\approx 10^7, d=28$), \falkon{} cannot scale beyond $100\mathrm{k}$ parameters without running out of memory (OOM) due to the quadratic storage requirements in $m$. 
With \algname{} we can keep the memory requirements fixed at the same level (essentially one $m_0\times m_0$ matrix, where $m_0=100\mathrm{k}$ for this experiment), and scale the number of parameters up to $3.2\mathrm{M}$, with a significant improvement in terms of test accuracy (see Table~\ref{tab:higgs-scale}).
Note that \algname{} does not face any inherent limitations to scaling $m$: memory remains fixed and the training time only scales up linearly. Furthermore, it is possible to keep the same $m$ and improve training times by reducing the block-size (see the ablation in \Cref{fig:ablation-inc-m0}).
In Figure~\ref{fig:higgs-scale} we also show how \joker{} can scale with the number of Random Fourier Features, noting that the improvements are not consistent across model sizes, and how \eigenpro{4} scales with the number of inducing points up to 500k centers, after which it goes OOM. Finally, \askotch{} solves the full KRR problem (and hence uses all possible $8.8\mathrm{M}$ parameters), but nevertheless does not outperform \algname{} either in time (\askotch{} requires \SI{4}{\hour} while \algname{} \SI{2}{\hour}) or in accuracy (see \Cref{tbl:large-scale}), likely due to the less accurate stochastic optimization algorithm used.

\paragraph{Inference times.}
Of course, using a larger amount of parameters affects the inference time. In \Cref{fig:inf-time} we compare the average (over 5 cross-validation folds) inference times on datasets of different sizes. 
Despite the linear increase in inference times with the number of parameters $m$, \algname{} is consistently the second fastest after \falkon{} (which always has a lower $m$). Both methods need a simple kernel-vector multiplication whose cost depends linearly on the number of parameters. \eigenpro{4} and \joker{} require batching to fit in memory which slows them down considerably. Meanwhile, \askotch{}'s inference time depends directly on the training-set size, hence it grows on larger datasets such as \emph{Higgs}, and becomes the main bottleneck e.g. on \emph{Taxi}, where we were unable to complete a single inference step on the full validation-set with about \num{200} million points in \SI{24}{\hour}. Overall, \algname{} outperforms the other algorithms considering the trade-off between accuracy (depicted in the last column of \Cref{fig:inf-time} as the relative rank averaged over all datasets) and inference time.

\begin{table}[t]
\centering
\caption{Medium/small-scale dataset comparisons. Cell colors as in \Cref{tbl:large-scale}; standard deviations in Appendix~\ref{apx:experiment_details}.}
\label{tbl:small-scale}
\small
\setlength{\tabcolsep}{4pt}
\begin{tabular*}{\textwidth}{@{\extracolsep{\fill}}l*{10}{c}@{}}
\toprule
 & \multicolumn{2}{c}{\makecell{MillionSongs\\[-1pt]{\footnotesize($n\approx10^5$)}}} & \multicolumn{2}{c}{\makecell{Benzene\\[-1pt]{\footnotesize($n\approx10^5$)}}} & \multicolumn{2}{c}{\makecell{MiniBooNE\\[-1pt]{\footnotesize($n\approx10^5$)}}} & \multicolumn{2}{c}{\makecell{FashionMNIST\\[-1pt]{\footnotesize($n\approx10^4$)}}} & \multicolumn{2}{c}{\makecell{SVHN\\[-1pt]{\footnotesize($n\approx10^4$)}}} \\
\cmidrule{2-3} \cmidrule{4-5} \cmidrule{6-7} \cmidrule{8-9} \cmidrule{10-11}
Algorithm & time & \makecell{rel. error \\{\footnotesize$\times 10^3$}} & time & \makecell{MSE\\{\footnotesize$\times 10^2$}} & time & AUC & time & accuracy & time & accuracy \\
\midrule
\falkon & {\boldmath \color{ForestGreen} $\SI{40}{\second}$} & ${4.492}$ & {\boldmath \color{ForestGreen} $\SI{27}{\second}$} & ${1.60}$ & {\boldmath \color{ForestGreen} $\SI{18}{\second}$} & {\color{YellowGreen} ${0.9733}$} & $\SI{30}{\second}$ & {\boldmath \color{ForestGreen} ${91.2\%}$} & $\SI{1}{\minute}$ & {\boldmath \color{ForestGreen} ${80.4\%}$} \\
\joker & $\SI{9}{\minute}$ & ${4.510}$ & $\SI{4}{\minute}$ & ${1.72}$ & $\SI{1}{\minute}$ & ${0.9717}$ & $\SI{32}{\second}$ & ${90.4\%}$ & {\boldmath \color{ForestGreen} $\SI{52}{\second}$} & ${76.7\%}$ \\
\askotch & $\SI{13}{\minute}$ & ${4.475}$ & $\SI{11}{\minute}$ & ${0.16}$ & $\SI{5}{\minute}$ & {\boldmath \color{ForestGreen} ${0.9741}$} & $\SI{1}{\hour}$ & {\color{YellowGreen} ${91.0\%}$} & $\SI{1}{\hour}$ & {\color{YellowGreen} ${79.9\%}$} \\
\eigenpro{4} & $\SI{19}{\minute}$ & ${4.502}$ & $\SI{7}{\minute}$ & ${1.99}$ & $\SI{5}{\minute}$ & ${0.9689}$ & $\SI{1}{\minute}$ & ${90.8\%}$ & $\SI{2}{\minute}$ & ${77.8\%}$ \\
\algname & $\SI{4}{\minute}$ & {\boldmath \color{ForestGreen} ${4.471}$} & $\SI{4}{\minute}$ & {\boldmath \color{ForestGreen} ${0.15}$} & $\SI{40}{\second}$ & {\color{YellowGreen} ${0.9718}$} & {\boldmath \color{ForestGreen} $\SI{14}{\second}$} & {\color{YellowGreen} ${91.2\%}$} & $\SI{1}{\minute}$ & {\color{YellowGreen} ${80.2\%}$} \\
\bottomrule
\end{tabular*}
\end{table}

For a broader comparison, we considered both large scale datasets (\Cref{tbl:large-scale}, training set sizes between \num{1} and \num{917} million points) and moderately sized datasets (\Cref{tbl:small-scale}, training set sizes between \num{50} and \num{500} thousand points). 
We ran all experiments on the same hardware to ensure reproducible results. 
When $n$ is large, we observe that \falkon{} always achieves the fastest training time. 
However, we observed that it is consistently outperformed by \joker{} and \askotch{} in terms of predictive accuracy: \falkon{} is constrained by its low maximum parameter count~\citep{eigenpro3} which cannot be increased due to its memory bottleneck inherent in the preconditioner.
\algname{} bypasses this bottleneck and indeed confirms that increasing the number of parameters can noticeably improve accuracy. This is especially true for large-scale datasets, where models with a low parameter count may fail to fully capture the complexity of the data.
The proposed model reaches state of the art accuracy among kernel-based methods on all the large scale datasets considered in \Cref{tbl:large-scale} while maintaining competitive training times compared to all other methods apart from \falkon{}.
Note that while each model has several hyperparameters that influence the final performance (most notably the kernel bandwidth), they were tuned independently to maximize accuracy while not exceeding reasonable training times. This latter constraint could not be fulfilled for \askotch{} on \emph{CIFAR10-5M} and for \eigenpro{4} on \emph{Taxi} due to the cost of training, and for \askotch{} on \emph{Taxi} due to the cost of inference. Interestingly, while the datasets and tasks remain the same, not all methods perform best with the same hyperparameters. For instance \askotch{} seems to prefer lower values for $\lambda$ compared to \algname{}, while \eigenpro{4} often worked best with different kernel bandwidths than the other methods.
Moving on to the medium-scale datasets shown in \Cref{tbl:small-scale}, we notice similar patterns: \falkon{} is often the fastest while \algname{} is often the most accurate. Note that on the \emph{Benzene} dataset \askotch{}, \falkon{} and \algname{} use a Mat{\'e}rn kernel which was not implemented in the libraries of \joker{} and \eigenpro{4}. Furthermore, \falkon{} was very unstable and this is reflected in the final accuracy. Since \askotch{} does not support multi-class classification, on the two image datasets \emph{FashionMNIST} and \emph{SVHN} we ran 10 one-vs-all models instead (leading to very high training times).
While on most of the other datasets \falkon{} had a lower total parameter count than the other algorithms, on \emph{FashionMNIST} and \emph{SVHN} we used the same $m$ for all models (except \askotch{}), and a smaller block size $m_0$ for \algname{}. 
This allowed us to observe that for smaller datasets, the fast \algname{} preconditioner allows to obtain the same accuracy with a faster training time.

\paragraph{Block size ablation.}
To expand on this last point we used two datasets (\emph{Airlines} and \emph{COMET MC}) to run an ablation on the time-accuracy tradeoff with different block sizes $m_0$. The results are shown in \Cref{fig:ablation-inc-m0}.
As expected, a larger block size lowers the test error and increases the training time. We note diminishing returns especially for \emph{Airlines}, and also note that the training time scales super-linearly with $m_0$.
The cost of the preconditioner explains this behavior. Factorizing a single block costs $\bigO(m_0^3)$, and with $q = m/m_0$ blocks the total preconditioning cost is $\bigO(m m_0^2)$, which is quadratic in $m_0$. This quadratic growth, however, is not directly visible in \Cref{fig:ablation-inc-m0}. When all blocks fit in memory simultaneously (hatched markers), the matrix-vector products with $\nysmat$ dominate the running time, and increasing $m_0$ has little effect on training time. Once the blocks become too large to be stored together (solid markers), they must be recomputed during training, and the preconditioning cost is no longer negligible compared to the matrix-vector products. This is visible as a sharp increase in marker area at the transition between the two regimes.
Together with the results of \Cref{fig:higgs-scale}, these experiments show that both the number of parameters $m$ and the block size $m_0$, can be used to improve predictive performance. Increasing $m$ raises the overall capacity of the model, while increasing $m_0$ improves the accuracy of the preconditioner, and consequently of the optimization procedure.

\section{Conclusions} \label{sec:conclusion}
In this work, we introduced \algname{}, a PCG-based solver for Nystr\"om-sketched KRR built around a novel sketched block Jacobi preconditioner whose memory cost, $\bigO(m_0^2)$, is independent of both the number of data $n$ and model size $m$. This decouples model size from memory while retaining the fast, step-size-free convergence of PCG. We provided a GPU-accelerated implementation and scaled \algname{} up to $3.2\mathrm{M}$ parameters with linear training time and consistently improving accuracy. Our results show that, under the same memory budget, \algname{} achieves higher accuracy than state-of-the-art methods with competitive training time.
Our work opens several research directions, including extending \algname{} to other loss functions, investigating different block partitioning strategies and incorporating coarse corrections that capture inter-block interactions~\citep{amg}. In particular, a logistic variant of \algname{} would enable kernel-based new physics searches and goodness-of-fit tests~\citep{Letizia2022,Grosso_2023} with substantially larger models than currently possible.

\FloatBarrier

\subsection*{Acknowledgments}
This work was partially supported by the European Commission (Horizon Europe grant ELIAS 101120237); and by the Ministry of Education, University and Research (FARE grant ML4IP R205T7J2KP); and by a France 2030 support managed by the Agence Nationale de la Recherche, under the reference ANR-23-PEIA-0004 (PDE-AI project). G. Vitale and B. Ayuso de Dios are members of the Gruppo Nazionale Calcolo Scientifico - Istituto Nazionale di Alta Matematica (GNCS-INdAM). We acknowledge the CINECA award under the ISCRA initiative, for the availability of high performance computing resources and support. This work represents only the view of the authors. The European Commission and the other organizations are not responsible for any use that may be made of the information it contains.

\subsection*{AI use statement}
In this work, we used generative AI tools for grammar and language polishing, text editing, and the development of utility scripts, including scripts for plotting and Bash scripts for launching experiments. We take responsibility for the final content of this work, including text, claims, or artifacts produced with the aid of generative AI.

\printbibliography

\newpage

\appendix

\section{Further comparison with the literature}
\label{apx:further_comparison}
As discussed in this paper, the scalability of kernel methods is fundamentally constrained by memory requirements. We overcame this bottleneck by developing \algname{}, an algorithm capable of scaling to large models and datasets. Other modern kernel methods address the issue of memory constraints in different ways. In this section, to the best of our knowledge, we provide a comprehensive overview of current state of the art methods and notable developments in large scale kernel learning. In the following, when a block $\mathcal{B}$ of $m$ indices is sampled from $\{1, \dots, n\}$, we denote by $I_{nm} \in \bbR^{n \times m}$ the submatrix of the $n \times n$ identity matrix formed by selecting all the $n$ rows and the $m$ columns indexed by $\mathcal{B}$. In particular, applying $I_{nm}^\top$ to a vector of $\bbR^n$ samples the components corresponding to $\mathcal{B}$, while applying $I_{nm}$ to a vector of $\bbR^m$ extends it to a vector of length $n$ by filling with zeros the components corresponding to the complement $\mathcal{B}_C = \{1, \dots, n\} \setminus \mathcal{B}$.

\paragraph{FALKON \citep{falkon_paper1, falkon_paper2}.} The \falkon{} algorithm consists of running PCG on problem \eqref{eq:NKRR_system} with preconditioner
\begin{equation}
\label{eq:flk_prec}
    \precond_{\text{FLK}} = \left( \frac{n}{m} \Kmm^2 + \lambda n \Kmm \right)^{-1}.
\end{equation}
We note that $\precond_{\text{FLK}}$ can be retrieved as a special case of the \algname{} preconditioner when $q = 1$. In~\citet{falkon_paper2}, the authors develop an extremely efficient library capable of fully exploiting the capabilities of GPU hardware and multiple GPUs. Furthermore, they also generalize \falkon{} to general self concordant losses, such as the logistic loss, designing a \textsc{Log}\falkon{} algorithm as well. Despite its efficiency, \falkon{}'s clear bottleneck lies in its preconditioner, which imposes a memory complexity of $\bigO(m^2)$. This bottleneck disallows \falkon{} to run with a large amount of Nystr\"om centers, typically running out of memory for values of $m$ around $10^5$.

\paragraph{RPCholesky, KRILL \citep{krill}.} In this paper the authors propose to solve both the full \eqref{eq:KRR_system} and Nystr\"om \eqref{eq:NKRR_system} problem via CG, designing a different preconditioner for the two. For the full system \eqref{eq:KRR_system}, when the amount of data is modest and the eigenvalues of $\Knn$ decay sufficiently rapidly, they propose the \rpchol{} preconditioner, generating a low-rank approximation of $\Knn$ via a randomly pivoted Cholesky decomposition~\citep{rpchol}. For the Nystr\"om system \eqref{eq:NKRR_system}, they propose the \krill{} preconditioner, defined as
\begin{equation}
\label{eq:krill_prec}
    \precond_{\text{KRL}} = \left( n (\Phi \Knm)^\top (\Phi \Knm) + \lambda n \Kmm \right)^{-1},
\end{equation}
where $\Phi \in \bbR^{m_0 \times n}$ a sparse random sign embedding~\citep{random_embeddings} in a suitable lower dimension $m_0$. We note that the \falkon{} preconditioner can be retrieved as a special case of \eqref{eq:krill_prec} when $m_0 = m$ and $\Phi = \frac{1}{\sqrt{m}}I_{nm}^\top$ samples the rows corresponding to the sampled Nystr\"om centers. By construction \krill{} incurs in the same memory bottleneck as \falkon{}.

\paragraph{EigenPro 4 \citep{eigenpro4}.} The latest entry in the \eigenpro{}family. The original \eigenpro{}~\citep{eigenpro1} solves problem \eqref{eq:KRR_system} on the full data in the interpolation regime, that is, with $\lambda = 0$. The authors consider a preconditioned gradient iteration
\begin{equation}
    \cv_{t+1} = \cv_t + \stepsize \precond_{\text{EP}} (\widehat{y} - \Knn \cv_t),
\end{equation}
where $\precond_{\text{EP}}$ is a preconditioner built to flatten the spectrum of $\Knn$. In particular, for a fixed number $m_0$ of eigenvalues, $\precond_{\text{EP}}$ is defined as
\begin{equation}
    \precond_{\text{EP}} = I - \sum_{i=1}^{m_0} \left( 1 - \damping \frac{\sigma_{m_0 + 1}}{\sigma_i} \right) u_i u_i^\top,
\end{equation}
where $\damping > 0$ is a damping parameter and $\sigma_i$ is the $i$-th largest eigenvalue of $\Knn$ with unit-norm eigenvector $u_i$. This approach presents two problems, namely the cost of computing the full gradient at each iteration and the cost of computing the exact leading eigenpairs of $\Knn$. The former problem is solved by implementing a stochastic gradient iteration in practice. For the latter issue the authors propose to compute the eigenpairs $(\sigma_i, u_i)$ approximately using the randomized SVD decomposition proposed by~\citet{rsvd}. \eigenpro{2}~\citep{eigenpro2} refactored the first \eigenpro{}to fully exploit modern GPU architectures. In particular the top eigenpairs are now approximated via a Nystr\"om based SVD method adapted from~\citet{nystrom}. This approach requires computing and storing a Nystr\"om approximation of $\Knn$ of rank at least equal to (and typically much larger than) $m_0 + 1$, incurring a memory footprint of $\Omega(m_0^2)$. As the size of datasets kept on growing, \eigenpro{3}~\citep{eigenpro3} was introduced. \eigenpro{3} disentangles the size of the data $n$ from the size of the model $m$, solving problem \eqref{eq:NKRR_system} via the projected preconditioned gradient iteration
\begin{equation}
    \cvn_{t+1} = \cvn_t + \stepsize \Kmm^{-1} \Knm^\top \precond_{\text{EP}} (\widehat{y} - \Knm \cvn_t).
\end{equation}
Again, stochastic gradient is employed in practice. This iteration introduces a memory bottleneck of $\bigO(m^2)$ due to the inversion of $\Kmm$. However, solving a system with $\Kmm$ can be seen as a new learning problem on the data $\{\widetilde{x}_1, \dots, \widetilde{x}_m\}$, which can be solved using \eigenpro{2}. \eigenpro{3} therefore approximates the projection step using \eigenpro{2}. Projecting at each iteration turned out to be too expensive in terms of time, therefore \eigenpro{4}~\citep{eigenpro4} finally introduced delayed projections, resulting in a variation of \eigenpro{3} where the projection is performed only periodically.
    
\paragraph{Joker \citep{joker}.} The authors argue that the main limitations of kernel methods are the memory requirements and the fact that research mainly focuses on KRR, while other methods lack studying. They introduce \joker{} (Joint Optimization Framework for Lightweight Kernel Machines), a general algorithm for different loss functions with low memory overhead. In the case of KRR, \joker{} aims to solve the dual problem
\begin{equation}
\label{eq:joker_dual_problem}
    \min_{\cv \in \bbR^n} \frac{1}{2} \cv^\top \Knn \cv - \widehat{y}^\top \cv + \frac{\lambda}{2} \| \cv \|^2.
\end{equation}
The chosen optimization method to solve \eqref{eq:joker_dual_problem} is block coordinate descent~\citep{block_coordinate_descent}. The full index set $\{1, \dots, n\}$ is partitioned into static, predetermined candidate blocks, and at each iteration $t$ a block $\mathcal{B} = \mathcal{B}(t)$ of cardinality $m_0$ is sampled uniformly at random. The block of coordinates $\cv_t^{\mathcal{B}}$ corresponding to $\mathcal{B}$ is then updated by solving
\begin{equation}
\label{eq:joker_dbcd}
    \min_{\cv^{\mathcal{B}} \in \bbR^{m_0}} \frac{1}{2} (\cv^{\mathcal{B}})^\top \Kmzmz \cv^{\mathcal{B}} + (\cv^{\mathcal{B}_C})^\top \Kcomplement \cv^{\mathcal{B}} + \frac{\lambda}{2} \| \cv^{\mathcal{B}} \|^2 - (\widehat{y}^{\mathcal{B}})^\top \cv^{\mathcal{B}},
\end{equation}
where $\Kmzmz = I^\top_{nm_0} \Knn I_{nm_0}$ is attained by sampling rows and columns corresponding to $\mathcal{B}$ from $\Knn$ and $\Kcomplement$ is attained by sampling rows corresponding to $\mathcal{B}_C$ and columns corresponding to $\mathcal{B}$ from $\Knn$. In turn, problem \eqref{eq:joker_dbcd} is solved using a trust region method~\citep{trust_region}. This leads to a constrained minimization of a quadratic form. For more general losses, there are also box constrained to be considered, and to deal with this problem the authors propose a heuristic truncated CG-Steihaug method~\citep{cg_steihaug}. Lastly, to alleviate computations, the authors also propose \textsc{Inexact} \joker{}, a variation of \joker{} incorporating an approximation of $\Knn$ using $m$ Random Fourier Features~\citep{rff}.
    
\paragraph{Skotch, ASkotch, SketchySAGA \citep{askotch}.} This paper originally proposed two methods to solve the full problem \eqref{eq:KRR_system} (\skotch{} and \askotch{}) and specialized the existing \sketchysaga{}~\citep{sketchysaga} to the Nystr\"om problem \eqref{eq:NKRR_system}. The most recent version eliminated the \sketchysaga{} approach to focus entirely on full KRR with \skotch{} (Scalable kernel optimization and training with coordinate sketch-and project and approximate Hessians) and \askotch{} (Accelerated \skotch{}). These are sketch-and-project (SAP) based methods~\citep{sap}. In particular, they implement an approximate version of the randomized Newton method~\citep{randomized_newton}, whose iteration is given by
\begin{equation}
    \cv_{t+1} = \cv_t - I_{n\widebar{m}} \precond_{\text{NWT}} ((K_{n\widebar{m}} + \lambda I_{n\widebar{m}})^\top \cv_t - I_{n\widebar{m}}^\top \widehat{y}),
\end{equation}
where at each iteration $t$ a block $\mathcal{B} = \mathcal{B}(t)$ of cardinality $\widebar{m}$ is sampled from $\{1, \dots, n\}$, $K_{n\widebar{m}} = \Knn I_{n\widebar{m}} \in \bbR^{n \times \widebar{m}}$, $\Kmm = I_{n\widebar{m}}^\top \Knn I_{n\widebar{m}} \in \bbR^{\widebar{m} \times \widebar{m}}$ and $\precond_{\text{NWT}} = (\Kmm + \lambda I)^{-1}$. The authors implement both uniform sampling and approximate leverage score sampling, the latter attained via the BLESS algorithm~\citep{bless}. The default suggested sampling parameter is $\widebar{m} = n / 100$, making the solution with $\precond_{\text{NWT}}$ at each step unfeasible in large scale scenarios. The authors therefore replace the matrix $\Kmm$ with a rank $m_0$ randomized Nystr\"om approximation, leading to the iteration
\begin{equation}
\label{eq:skotch_iteration}
    \cv_{t+1} = \cv_t - \stepsize I_{n\widebar{m}} (\Kmm_{\text{nys}} + \damping I)^{-1} ((K_{n\widebar{m}} + \lambda I_{n\widebar{m}})^\top \cv_t - I_{n\widebar{m}}^\top \widehat{y}),
\end{equation}
where $\Kmm_{\text{nys}} = (\Kmm \Omega)(\Omega^\top \Kmm \Omega)^{-1} (\Kmm \Omega)^\top \in \bbR^{\widebar{m} \times \widebar{m}}$ with $\Omega \in \bbR^{\widebar{m} \times m_0}$ a Gaussian random matrix, $\damping > 0$ is a damping parameter and $\stepsize$ is a suitable step size. The authors propose a default damping parameter dependent on $\lambda$. Finally, the step size $\stepsize$ is automatically selected at each iteration to be the reciprocal of a preconditioned smoothness constant, namely
\begin{equation}
\label{eq:skotch_stepsize}
    \stepsize = \frac{1}{\sigma_{\max}\left((\Kmm_{\text{nys}} + \damping I)^{-1/2} (\Kmm + \lambda I) (\Kmm_{\text{nys}} + \damping I)^{-1/2} \right)},
\end{equation}
computed via randomized powering~\citep{randomized_powering}. Iteration \eqref{eq:skotch_iteration} with step size \eqref{eq:skotch_stepsize} defines the \skotch{} algorithm. Finally, by incorporating Nesterov acceleration~\citep{nesterov}, the authors introduce \askotch{}, an accelerated variant that constitutes their core algorithmic contribution.
\begin{table}[ht]
    \centering
    \caption{Parameters and memory requirements of state of the art algorithms.}
    \label{tab:algorithm_comparison}
    \begin{tabular}{lcccc}
    \toprule
    Algorithm & $n$ & $m$ & $m_0$ & Memory \\
    \midrule
    \falkon{} & data size & model size (Nystr\"om) & \na & $\bigO(m^2)$ \\
    \krill{} & data size & model size (Nystr\"om) & embedding dimension & $\bigO(m^2)$ \\
    \eigenpro{4} & data size & model size (Nystr\"om) & preconditioner rank & $\Omega(m_0^2)$ \\
    \joker{} & data size & model size (RFF) & block size & $\bigO(m_0^2)$ \\
    \askotch{} & data and model size & \na{} (Full KRR) & rank of Nys. approx. & $\bigO(m_0^2)$ \\
    \algname{} & data size & model size (Nystr\"om) & block size & $\bigO(m_0^2)$ \\
    \bottomrule
    \end{tabular}
\end{table}

\section{Block preconditioning strategies and comparisons}
\label{apx:block_prec}
In Section~\ref{sec:algorithm} we mentioned that the \algname{} preconditioner can be thought as a sketched version of the \bj{} preconditioner. Formally, consider a block diagonal partition of $\nysmat$ in $q$ blocks of size $m_0$, and the corresponding partition of the Nystr\"om data \eqref{eq:data_partition}. For $j = 1, \dots, q$, the $j$-th diagonal block of $\nysmat$ is
\begin{equation}
    \nysmat_j = \KnmjT \Knmj + \lambda n \Kmjmj,
\end{equation}
where $(\Knmj)_{il_j} = \kernel(x_i, \widetilde{x}_{l_j})$ for $i = 1, \dots, n$ and $l_j = (j - 1) m_0 + 1, \dots, j m_0$. The standard \bj{} preconditioner is
\begin{equation}
\label{eq:bj_prec}
    \precond_{\text{BJ}} = \operatorname{diag} \left(\nysmat_1^{-1}, \dots, \nysmat_q^{-1} \right).
\end{equation}
In principle, preconditioner \eqref{eq:bj_prec} already alleviates the memory bottleneck, since the size of the blocks does not depend on $m$. However, each block $\nysmat_j$ has the same structure as the original matrix $\nysmat$, where dimensionally $m_0$ replaces $m$. This implies that computing and inverting the blocks $\nysmat_j$ presents the same problems that computing and inverting $\nysmat$ presented before, namely the numerical instability and $\bigO(nm_0^2)$ cost of computing the Gram products $\KnmjT \Knmj$. To solve this issue, we may sketch each block $\nysmat_j$ by sampling from $\Knmj$ the rows corresponding to the $j$-th block. After normalization, the result of this sketching procedure is precisely the block $\nysmatnys_j$, as illustrated in Figure~\ref{fig:first_derivation}, and inverting and stacking these sketched blocks diagonally yields exactly the \algname{} preconditioner~\eqref{eq:blk_prec}.

In Appendix~\ref{apx:further_comparison} we noted that the \falkon{} preconditioner $\precond_\text{FLK}$ \eqref{eq:flk_prec} is a special case of the \algname{} preconditioner $\precond$ \eqref{eq:blk_prec} when $q = 1$. Following this point of view, we could also think of $\precond$ as an approximation of $\precond_\text{FLK}$ to be employed when $m$ is too large. In  particular, let $\nysmatnys' = \frac{n}{m} \Kmm^2 + \lambda n \Kmm$, so that \falkon{}'s preconditioner reads as $\precond_{\text{FLK}} = (\nysmatnys')^{-1}$. We could perform a block diagonal approximation of $\nysmatnys'$ rather than $\nysmat$, creating a hybrid \bjf{} preconditioner. For $j = 1, \dots, q$, let $\nysmatnys'_j$ be the $j$-th diagonal block of $\nysmatnys'$, that is
\begin{equation}
    \nysmatnys'_j = \frac{n}{m} \KmmjT \Kmmj + \lambda n \Kmjmj,
\end{equation}
where $\Kmmj \in \bbR^{m \times m_0}$ is defined by $(\Kmmj)_{i{l_j}} = \kernel(\widetilde{x}_i, \widetilde{x}_{l_j})$ for $i = 1, \dots, m$ and $l_j = (j - 1) m_0 + 1, \dots, j m_0$. We define the \bjf{} preconditioner as
\begin{equation}
\label{eq:bjf_prec}
    \precond_{\text{BJF}} = \operatorname{diag}\left( (\nysmatnys'_1)^{-1}, \dots, (\nysmatnys'_q)^{-1} \right).
\end{equation}
Again, preconditioner \eqref{eq:bjf_prec} does not present any memory issues, but remains untractable in regimes where $m$ is very large, due to the instability and $\bigO(mm_0^2)$ cost of forming the Gram product $\KmmjT \Kmmj$.
We may again approximate the blocks $\nysmatnys'_j$ by sketching the rows of $\Kmmj$, as illustrated in Figure~\ref{fig:second_derivation}. This procedure produces again the approximation $\nysmatnys_j = \frac{nq}{m_0} \Kmjmj^2 + \lambda n q \Kmjmj$, yielding once more the \algname{} preconditioner \eqref{eq:blk_prec}.
\begin{figure}[ht]
    \centering
    \colorlet{blockgray}{gray!30}
    \begin{tikzpicture}[scale=1,
        seconddiag/.style={thick, dashed}
    ]
    \def\bs{1.2}
    \def\ms{2.4}

    \def\xMtwo{3.2}
    \def\xMthree{5.8} 
    \def\xMfour{9.0}

    \def\yLineTwo{-4.2}
    \def\yLineTwoMid{-3.0} 

    \begin{scope}[shift={(0,0)}]
        \fill[blockgray] (0,0) rectangle (\ms,\ms);
        \fill[malgared] (0,\bs) rectangle (\bs,\ms);
        
        \draw[seconddiag] (\bs,0) -- (\bs,\bs) -- (\ms,\bs);

        \draw[thick] (0,0) rectangle (\ms,\ms);
        \draw[thick] (0,\bs) rectangle (\bs,\ms);

        \node at (\bs/2, \bs + \bs/2) {$\propto \nysmatnys'_1$};
        
        \draw[decorate,decoration={brace,amplitude=5pt},thick] 
            (0,\ms+0.05) -- (\bs,\ms+0.05) node[midway, above=6pt] {$m_0$};
        \draw[decorate,decoration={brace,amplitude=5pt},thick] 
            (-0.05,\bs) -- (-0.05,\ms) node[midway, left=6pt] {$m_0$};
            
        \node[below=8pt] at (\bs, 0) {$\nysmatnys' \propto \Kmm^2 + \Kmm$};
    \end{scope}

    \node at (2.8, \bs) {\Large $=$};

    \begin{scope}[shift={(\xMtwo,0)}]
        \fill[blockgray] (0,0) rectangle (\ms,\ms);
        \fill[malgared] (0,\bs) rectangle (\ms,\ms);
        
        \draw[thick] (0,0) rectangle (\ms,\ms);
        \draw[thick] (0,\bs) rectangle (\ms,\ms);
        
        \node at (\bs, \bs + \bs/2) {$K_{mm_0}^{(1)\top}$};
    \end{scope}

    \begin{scope}[shift={(\xMthree,0)}]
        \fill[blockgray] (0,0) rectangle (\ms,\ms);
        \fill[malgared] (0,0) rectangle (\bs,\ms);
        
        \draw[thick] (0,0) rectangle (\ms,\ms);
        \draw[thick] (0,0) rectangle (\bs,\ms);
        
        \node at (\bs/2, \bs) {$K_{mm_0}^{(1)}$};
    \end{scope}

    \node at (8.6, \bs) {\Large $+$};

    \begin{scope}[shift={(\xMfour,0)}]
        \fill[blockgray] (0,0) rectangle (\ms,\ms);
        \fill[malgared] (0,\bs) rectangle (\bs,\ms);
        
        \draw[seconddiag] (\bs,0) -- (\bs,\bs) -- (\ms,\bs);

        \draw[thick] (0,0) rectangle (\ms,\ms);
        \draw[thick] (0,\bs) rectangle (\bs,\ms);
        
        \node at (\bs/2, \bs + \bs/2) {$\Kmm_1$};
    \end{scope}

    \node at (2.8, \yLineTwoMid) {\Large $\approx$};

    \begin{scope}[shift={(\xMtwo,\yLineTwo)}]
        \fill[blockgray] (0,0) rectangle (\ms,\ms);
        \fill[malgared] (0,\bs) rectangle (\bs,\ms);
        
        \draw[seconddiag] (\bs,\bs) -- (\ms,\bs);

        \draw[thick] (0,0) rectangle (\ms,\ms);
        \draw[thick] (0,\bs) rectangle (\bs,\ms);
        
        \node at (\bs/2, \bs + \bs/2) {$\Kmm_1$};
    \end{scope}

    \begin{scope}[shift={(\xMthree,\yLineTwo)}]
        \fill[blockgray] (0,0) rectangle (\ms,\ms);
        \fill[malgared] (0,\bs) rectangle (\bs,\ms);
        
        \draw[seconddiag] (\bs,0) -- (\bs,\bs);

        \draw[thick] (0,0) rectangle (\ms,\ms);
        \draw[thick] (0,\bs) rectangle (\bs,\ms);
        
        \node at (\bs/2, \bs + \bs/2) {$\Kmm_1$};
    \end{scope}

    \node at (8.6, \yLineTwoMid) {\Large $+$};

    \begin{scope}[shift={(\xMfour,\yLineTwo)}]
        \fill[blockgray] (0,0) rectangle (\ms,\ms);
        \fill[malgared] (0,\bs) rectangle (\bs,\ms);
        
        \draw[seconddiag] (\bs,0) -- (\bs,\bs) -- (\ms,\bs);

        \draw[thick] (0,0) rectangle (\ms,\ms);
        \draw[thick] (0,\bs) rectangle (\bs,\ms);
        
        \node at (\bs/2, \bs + \bs/2) {$\Kmm_1$};
    \end{scope}

    \end{tikzpicture}
    \caption{First block of $\nysmatnys'$ (above), and sketched approximation (below). Constants omitted for visual clarity.}
    \label{fig:second_derivation}
\end{figure}

Overall, we have introduced three block preconditioners, namely \bj{} \eqref{eq:bj_prec}, \bjf{} \eqref{eq:bjf_prec} and \algname{} \eqref{eq:blk_prec}. Despite them sharing the same exact memory complexity, their key differences lie in the information carried by each block and the cost to solve with them, as detailed in Table~\ref{tab:block_preconditioners}.
\begin{table}[ht]
    \small
    \centering
    \caption{Difference in information and time cost between block preconditioners. FLOPS refers to the number of operations required to compute and invert a single block. The total cost of the preconditioner is given by multiplying this cost by the number of blocks $q = m/m_0$.}
    \label{tab:block_preconditioners}
    \begin{tabular}{llll}
    \toprule
    \textbf{Preconditioner} & \textbf{$j$-th block} & \textbf{FLOPS} & \textbf{Interactions captured by the $j$-th block} \\ \midrule
    $\precond_{\text{BJ}}$ \eqref{eq:bj_prec} & $\left( \KnmjT \Knmj + \lambda n \Kmjmj \right)^{-1}$ & $\bigO(nm_0^2 + m_0^3)$ & $\{x_1, \dots, x_n\}$ and $\{\widetilde{x}_{(j - 1) m_0 + 1}, \dots, \widetilde{x}_{m_{jm_0}}\}$ \\ \addlinespace
    $\precond_{\text{BJF}}$ \eqref{eq:bjf_prec} & $\left( \frac{n}{m} \KmmjT \Kmmj + \lambda n \Kmjmj \right)^{-1}$ & $\bigO(mm_0^2 + m_0^3)$ & $\{\widetilde{x}_1, \dots, \widetilde{x}_m\}$ and $\{\widetilde{x}_{(j - 1) m_0 + 1}, \dots, \widetilde{x}_{m_{jm_0}}\}$ \\ \addlinespace
    $\precond$ \eqref{eq:blk_prec} & $\left( \frac{nq}{m_0}\Kmjmj^2 + \lambda n q \Kmjmj \right)^{-1}$ & $\bigO(m_0^3)$ & $\{\widetilde{x}_1^{(j)}, \dots, \widetilde{x}_{m_{j}}^{(j)}\}$ and $\{\widetilde{x}_{(j - 1) m_0 + 1}, \dots, \widetilde{x}_{m_{jm_0}}\}$ \\ \bottomrule
    \end{tabular}
\end{table}
Intuitively, standard \bj{} should yield superior performance compared to \bjf{}, which in turn should perform better than \algname{}. We show empirically that this expected gap in performance is marginal, and that performances at test time are fundamentally the same, making \algname{}, which is the only computationally viable preconditioner for large scale applications, a highly effective choice in practice. We conduct experiments on smaller datasets where the system in \eqref{eq:NKRR_system} can be solved exactly, enabling us to monitor the exact error at each iteration and directly compare the three preconditioning strategies. In particular, we consider three different metrics to evaluate the quality of the produced solutions over the number of iterations. Let $\cvn_t$ be the solution computed after $t$ iterations of left preconditioned CG with any preconditioner $\precond$ on problem \eqref{eq:NKRR_system}.

\paragraph{True relative error (matrix norm).} Let $\cvne$ be the exact solution to \eqref{eq:NKRR_system}, define
\begin{equation}
    \relerrmat(\cvn_t) = \frac{\|\cvne - \cvn_t\|_\nysmat}{\|\cvne\|_\nysmat}.
\end{equation}
By construction of the CG algorithm~\citep{cg_original}, $\relerrmat(\cvn_t)$ decreases monotonically with the number of iterations $t$.

\paragraph{Relative residual (preconditioner norm).} Define
\begin{equation}
    \relresprec(\cvn_t) = \frac{\|\Knm^\top \widehat{y} - \nysmat \cvn_t\|_\precond}{\|\Knm^\top \widehat{y}\|_\precond}.
\end{equation}
The relative residual can be measured in practice even if the exact solution $\cvne$ is not available. The quantity $\relresprec(\cvn_t)$ is not guaranteed to monotonically decrease, however if $\precond$ is a good preconditioner the behavior should be similar to that of the true relative error. In the extreme case when $\precond = \nysmat^{-1}$ it is immediate to check that $\relres_{\nysmat^{-1}}(\cvn_t) = \relerrmat(\cvn_t)$. We can think of the relative error and the relative residual as a measure of the training error.

\paragraph{Test error.} The error of the model computed at iteration $t$ on the test set. From the machine learning point of view this is the quantity of most interest. The exact metric depends on the dataset considered and on the task performed (classification, regression).

\vspace{\baselineskip}
\noindent
Given that we are bounded to run these experiments on smaller scales, we implemented a basic working version of \bj{} and \bjf{},\footnote{With an abuse of language, we call \bj{} and \bjf{} both the preconditioners and the corresponding left PCG algorithms.} explicitly forming and storing all kernel matrices. To have a fair comparison, we re-implemented \algname{} in the same way. We note that in this scenario we could store all blocks explicitly and solve with them in parallel; however, to simulate the constraints of large scale scenarios, we still recompute them from scratch at each iteration. For this same reason, even though we form the matrix $\nysmat$ explicitly to compute the exact solution $\cvne$, we still implement the matrix-vector product with $\nysmat$ using kernel matrices as in line \ref{line:matvect} of Algorithm~\ref{alg:blk_algorithm_long}. We ran experiments across different datasets and consistently observed the same results: at train time, \bj{} performs better than \bjf{}, which in turn tends to perform better than \algname{}, though at the cost of much longer training times. However, at test time, the performances of the three methods are almost indistinguishable. We believe this phenomenon might be partly caused by additional regularization introduced by the approximation of the blocks, and we leave a rigorous theoretical analysis for future work. In particular, this implies that the information lost from the standard \bj{} preconditioner might be compensated for by appropriately choosing the regularization parameter $\lambda$. For the sake of compactness, we report results for only two datasets, obtained by subsampling $100\ 000$ points from the HIGGS and AIRLINES datasets and performing a $5$-fold cross validation, effectively resulting in $n = 80\ 000$ training points. We run these experiments in double precision to minimize the impact of numerical errors, since we have to compute the exact solution $\cvne$ and invert huge Gram matrices for the \bj{} preconditioner, as mentioned in Table~\ref{tab:block_preconditioners}.
\begin{table}[ht]
    \centering
        \caption{Comparison of different block preconditioning strategies on subsamples of the HIGGS and AIRLINES datasets. Means over $5$-fold cross validation, standard deviations omitted for space.}
    \label{tab:subsampled_higgs_airlines}
    \small
    \begin{tabular}{l ccc ccc}
    \toprule
    & \multicolumn{3}{c}{HIGGS} & \multicolumn{3}{c}{AIRLINES} \\
    \cmidrule(lr){2-4} \cmidrule(lr){5-7}
     & $1-\mathrm{AUC}$ & time (s) & time w.r.t. \algname{}
     & MSE & time (s) & time w.r.t. \algname{} \\
    \midrule
    \bj{}      & $0.2358$ & $103.17$ & $\times 12.98$
               & $0.9627$ & $87.36$ & $\times 15.29$ \\
    \bjf{}     & $0.2349$ & $17.56$ & $\times 2.21$
               & $0.9629$ & $6.35$ & $\times 1.11$ \\
    \algname{} & $0.2335$ & $7.95$ & $\times 1.00$
               & $0.9629$ & $5.72$ & $\times 1.00$ \\
    \bottomrule
    \end{tabular}
\end{table}

\begin{figure}[ht]
  \centering
  \setlength{\abovecaptionskip}{6pt}
  \includegraphics[width=\textwidth]{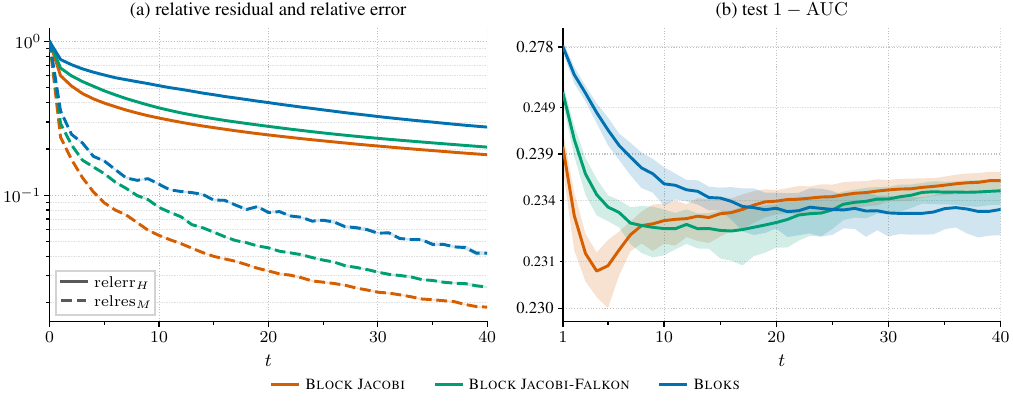}%
  \caption{Subsample of the HIGGS dataset, $5$-fold cross-validation. (a) relative error $\relerrmat$ (solid) and relative residual $\relresprec$ (dashed); (b) test $1-\mathrm{AUC}$. Curves are averages over the folds, shaded bands are the standard deviations.}
  \label{fig:relerr_relres_loss_higgs_combined}
\end{figure}

\paragraph{Subsampled HIGGS (Table~\ref{tab:subsampled_higgs_airlines}, Figure~\ref{fig:relerr_relres_loss_higgs_combined}).}
For all three methods we sample $m = 10^4$ Nystr\"om centers and use blocks of size
$m_0 = 10^3$. All other parameters are those of the experiments in
Section~\ref{sec:experiments}, reported in Table~\ref{tab:parameters_details}, except
that we run $t_{\max} = 40$ iterations in order to obtain longer convergence curves.
Test performances are comparable across the three methods, and as expected \algname{} is the fastest, with \bjf{} taking more than twice as long and plain \bj{} almost thirteen times as long. Panel (b) of Figure~\ref{fig:relerr_relres_loss_higgs_combined} is particularly informative. The test $1-\mathrm{AUC}$ of \bj{} drops sharply within the first few iterations and then increases again as the iterations proceed, a clear sign of overfitting. The same behaviour occurs for \bjf{}, though less markedly, whereas the \algname{} curve is almost monotone. This suggests that approximating the blocks acts as an additional source of regularization. Consequently, after $40$ iterations, \algname{} has not overfitted and attains the best test performance.

\begin{figure}[ht]
  \centering
  \setlength{\abovecaptionskip}{6pt}
  \includegraphics[width=\textwidth]{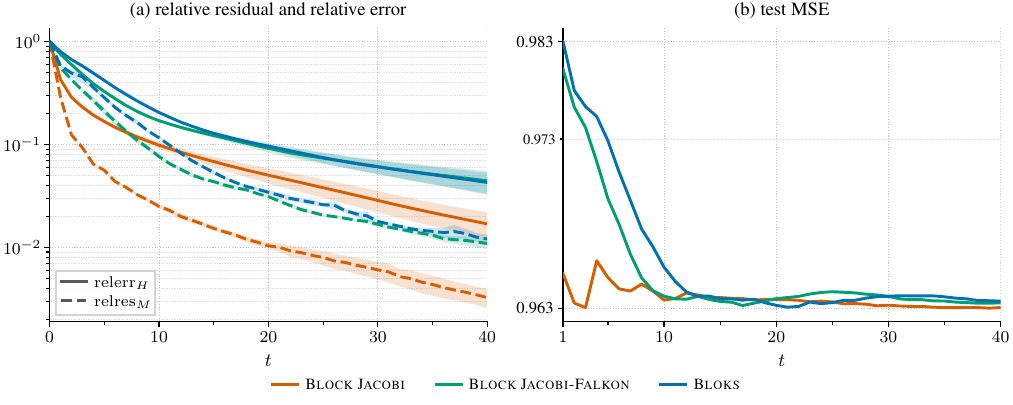}%
  \caption{Subsample of the AIRLINES dataset, $5$-fold cross-validation.
    (a) relative error $\mathrm{relerr}_H$ (solid) and relative residual
    $\mathrm{relres}_M$ (dashed); (b) test MSE. All curves are averages over the folds. The shaded bands in~(a) are the standard deviations across the folds; in~(b) we do not report them, as they are of order $4 \times 10^{-2}$, much larger than the gaps between the curves, and would overlap into a single transparent block, collapsing the curves in a single flat line.}
  \label{fig:relerr_relres_loss_airlines_combined}
\end{figure}

\paragraph{Subsampled AIRLINES (Table~\ref{tab:subsampled_higgs_airlines}, Figure~\ref{fig:relerr_relres_loss_airlines_combined}).} For all three methods we sample $m = 4 \times 10^3$ Nystr\"om centers and use blocks of size $m_0 = 2 \times 10^3$. All other parameters are those of the experiments in Section~\ref{sec:experiments}, reported in Table~\ref{tab:parameters_details}, except that again we run $t_{\max} = 40$ iterations and we lower the bandwidth of the kernel to $\sigma = 0.5$, as a larger value would cause the \bj{} blocks to be numerically singular. Test performances are now indistinguishable, in agreement with panel (b) of Figure~\ref{fig:relerr_relres_loss_airlines_combined}, and, in contrast with the HIGGS experiment, no method exhibits overfitting within $40$ iterations. Furthermore, we note that $n$ is the same as in the previous experiment on HIGGS, whereas $m_0$ is larger and $m$ smaller. According to Table~\ref{tab:block_preconditioners}, increasing $m_0$ should worsen the times of all three methods, since the cost per block grows at least quadratically in $m_0$ while the number of blocks $q = m/m_0$ only decreases linearly. The most penalized in absolute terms is \bj{}, due to the term $nm_0^2$ per block depending on the number of points $n$, the largest quantity involved, while the most penalized in relative terms is \algname{}, the only one whose preconditioning cost per block is driven purely by $m_0$. Decreasing $m$, on the other hand, benefits \bjf{} the most, as it is the only one whose preconditioning cost per block depends on $m$, although all three methods speed up, since each iteration requires a matrix-vector product with $\nysmat$ costing $\bigO(nm + m^2)$. This is also the reason why the times for AIRLINES are lower than the times for HIGGS. The combination of the two effects is what we observe in Table~\ref{tab:subsampled_higgs_airlines}, where \bjf{} becomes almost as fast as \algname{}, whereas \bj{} remains more than an order of magnitude slower. Note that in this experiment $m_0 = m/2$. In the extreme case where $m_0 = m$, \algname{} coincides with \bjf{} (and they both coincide with \falkon{}), and \bj{} becomes the exact solver $\nysmat^{-1}$.

\section{Bilateral vs. left preconditioning}
\label{apx:bilateral_vs_left_prec}
Problem \eqref{eq:NKRR_system} could be preconditioned using two distinct approaches, hereafter referred to as bilateral and left preconditioning (right preconditioning is also possible, but we do not consider it). Although mathematically equivalent, the former, also known as split preconditioning, is typically adopted for theoretical analysis, given that it requires an explicit symmetric decomposition of the preconditioner, whereas the latter is preferred in practical implementations, requiring only the action of the whole preconditioner on vectors, potentially as a black box. While our construction provides an explicit symmetric decomposition of the \algname{} preconditioner, making the bilateral formulation the natural candidate, the left variant proves significantly more efficient in practice. Indeed, thanks to the block diagonal structure, the \algname{} preconditioner can be decomposed as
\begin{equation}
    \precond = \preccf \preccf^\top, \qquad \preccf = \operatorname{diag}(\preccf_1, \dots, \preccf_q),
\end{equation}
where the blocks $\preccf_j$ are defined in \eqref{eq:blk_prec_decomposition}. We could precondition problem \eqref{eq:NKRR_system} by applying the factors of $\precond$ to the left and right of $\nysmat$, to preserve symmetry and be able to run CG. This approach leads to the preconditioned problem \eqref{eq:NKRR_system_prec}, and running CG on \eqref{eq:NKRR_system_prec} is a form of Preconditioned CG, which we may call bilateral PCG. Alternatively, if $\precond$ is accessed only through its action on vectors, even if formulation \eqref{eq:NKRR_system_prec_left} is not symmetric in general, PCG can be expressed via a change of variables, or equivalently a change of inner product~\citep{saad}, using solely the action of $\precond$ as a whole, leading to what we may call left PCG. Algorithm~\ref{alg:BPCG} and Algorithm~\ref{alg:LPCG} report, respectively, the bilateral and left variant of the PCG algorithm, assuming a total number $t_{\max}$ of iterations and starting point $\cvn_0 = 0$.
\begin{table}[ht]
\centering
\begin{minipage}[t]{0.48\textwidth}
\begin{algorithm}[H]
\caption{Bilateral PCG for \eqref{eq:NKRR_system} (CG for \eqref{eq:NKRR_system_prec})}
\label{alg:BPCG}
\begin{algorithmic}[1]
    \Statex \textbf{Input}: $\nysmat, \Knm^\top \widehat{y}, \preccf, t_{\max}$
    \Statex \textbf{Output}: $\cvn \approx \nysmat^{-1}\Knm^\top \widehat{y}$
    \smallskip
    \State $r_0 = \preccf^\top \Knm^\top \widehat{y}$
    \State $p_0 = r_0$
    \For{$t = 0$ to $t_{\max}-1$}
        \State $\alpha_t = \frac{r_t^\top r_t}{p_t^\top \precmatlr p_t}$
        \State $\cvnp_{t+1} = \cvnp_t + \alpha_t p_t$
        \State $r_{t+1} = r_t - \alpha_t \precmatlr p_t$
        \State $\beta_t = \frac{r_{t+1}^\top r_{t+1}}{r_t^\top r_t}$
        \State $p_{t+1} = r_{t+1} + \beta_t p_t$
    \EndFor
    \State \Return $\cvn = \preccf \cvnp_{t_{\max}}$
\end{algorithmic}
\end{algorithm}
\end{minipage}
\hfill
\begin{minipage}[t]{0.48\textwidth}
\begin{algorithm}[H]
\caption{Left PCG for \eqref{eq:NKRR_system}}
\label{alg:LPCG}
\begin{algorithmic}[1]
    \Statex \textbf{Input}: $\nysmat, \Knm^\top \widehat{y}, \precond, t_{\max}$
    \Statex \textbf{Output}: $\cvn \approx \nysmat^{-1}\Knm^\top \widehat{y}$
    \smallskip
    \State $r_0 = \Knm^\top \widehat{y}$
    \State $p_0 = \precond r_0$
    \For{$t = 0$ to $t_{\max}-1$}
        \State $\alpha_t = \frac{r_t^\top \precond r_t}{p_t^\top \nysmat p_t}$
        \State $\cvn_{t+1} = \cvn_t + \alpha_t p_t$
        \State $r_{t+1} = r_t - \alpha_t \nysmat p_t$
        \State $\beta_t = \frac{r_{t+1}^\top \precond r_{t+1}}{r_t^\top \precond r_t}$
        \State $p_{t+1} = \precond r_{t+1} + \beta_t p_t$
    \EndFor
    \State \Return $\cvn = \cvn_{t_{\max}}$
\end{algorithmic}
\end{algorithm}
\end{minipage}
\end{table}

In principle, the number of operations performed by both algorithms is the same. For each iteration $t$, in the bilateral variant, the action of the preconditioned matrix $\precmatlr$ is used only once to compute the quantity $\precmatlr p_t$. In the left variant the action of the matrix is decoupled from that of the preconditioner, but both are used only once: the action of the matrix to compute the product $\nysmat p_t$, and the preconditioner to compute $\precond r_{t+1}$ (note that $\precond r_t$ is available from the previous iteration, and $\precond r_0$ is computed outside the loop at the beginning). Even outside the loop, the bilateral variant uses the factor $\preccf^\top$ at the beginning to transform the right hand side, and the factor $\preccf$ to transform the solution back, whereas in the left variant the whole $\precond$ is used once at the beginning to compute the first preconditioned residual. Therefore, if the preconditioner $\precond = \preccf \preccf^\top$ is explicitly stored in memory, both algorithms have the same exact computational cost. However, due to memory constraints, the \algname{} preconditioner computes each block on the fly at each iteration. This makes the left variant far more convenient, as computing the action of $\precmatlr$ requires forming the blocks twice, before and after the multiplication by $\nysmat$, as detailed in Algorithm~\ref{alg:blk_lr}, whereas the left variant, solving with $\precond$ as a whole, requires materializing each block only once per iteration, as detailed in Algorithm~\ref{alg:blk_l}. Indeed, since the blocks do not overlap, they act on independent parts of the vector they are applied to.
\begin{table}[ht]
\centering
\begin{minipage}[t]{0.48\textwidth}
\begin{algorithm}[H]
\caption{\algname{} preconditioner - bilateral}
\label{alg:blk_lr}
\begin{algorithmic}[1]
    \Statex \textbf{Input}: $p \in \bbR^m; \Kmm, m_0$
    \Statex \textbf{Output}: $\preccf^\top \nysmat \preccf p \in \bbR^m$
    \smallskip
    \State $q = m / m_0$
    \For{$j = 1$ to $q$}
        \State $T_j = \chol(\Kmjmj)$
        \State $A_j = \chol \left( \frac{1}{m_0}T_j T_j^\top + \lambda I \right)$
        \State $p^{(j)} \gets T_j^{-1}(A_j^{-1} p^{(j)})$
        \State \textbf{delete} $T_j, A_j$ \Comment{Free memory for next block}
    \EndFor
    \State $p \gets \nysmat p$
    \For{$j = 1$ to $q$}
        \State $T_j = \chol(\Kmjmj)$
        \State $A_j = \chol \left( \frac{1}{m_0}T_j T_j^\top + \lambda I \right)$
        \State $p^{(j)} \gets A_j^{-T}(T_j^{-T} p^{(j)})$
        \State \textbf{delete} $T_j, A_j$ \Comment{Free memory for next block}
    \EndFor
    \State \Return $p \gets \frac{1}{nq} p$
\end{algorithmic}
\end{algorithm}
\end{minipage}
\hfill
\begin{minipage}[t]{0.48\textwidth}
\begin{algorithm}[H]
\caption{\algname{} preconditioner - left}
\label{alg:blk_l}
\begin{algorithmic}[1]
    \Statex \textbf{Input}: $r \in \bbR^m; \Kmm, m_0$
    \Statex \textbf{Output}: $\precond r \in \bbR^m$
    \State $q = m / m_0$
    \For{$j = 1$ to $q$}
        \State $T_j = \chol(\Kmjmj)$
        \State $A_j = \chol \left( \frac{1}{m_0}T_j T_j^\top + \lambda I \right)$
        \State $r^{(j)} \gets T_j^{-1}(A_j^{-1}(A_j^{-T}(T_j^{-T} r^{(j)})))$
        \State \textbf{delete} $T_j, A_j$ \Comment{Free memory for next block}
    \EndFor
    \State \Return $r \gets \frac{1}{nq} r$
\end{algorithmic}
\end{algorithm}
\end{minipage}
\end{table}

\subsection{Comparison with FALKON and KRILL}
As noted in Appendix~\ref{apx:further_comparison}, the \falkon{} preconditioner $\precond_{\text{FLK}}$ \eqref{eq:flk_prec} is retrieved as a special case of the \algname{} preconditioner when $q = 1$. In particular, $\precond_{\text{FLK}}$ can be decomposed as
\begin{equation}
\label{eq:flk_prec_decomposition}
    \precond_{\text{FLK}} = \preccf_{\text{FLK}} \preccf_{\text{FLK}}^\top, \qquad \preccf_{\text{FLK}} = \frac{1}{\sqrt{n}} T^{-1} A^{-1}, \qquad T = \chol(\Kmm), \qquad A = \chol \left( \frac{1}{m}T T^\top + \lambda I \right),
\end{equation}
and the single block can be explicitly stored in memory. This implies that for \falkon{} there is no real advantage in using left preconditioning. Notably, for \falkon{} it is actually more convenient to run the bilateral variant of PCG. Indeed, the bilateral preconditioned matrix can be written as
\begin{equation}
\label{eq:flk_prec_mat}
    \preccf_{\text{FLK}}^\top \nysmat \preccf_{\text{FLK}} = A^{-T} \left( \frac{1}{n} T^{-T} \Knm^\top \Knm T^{-1} + \lambda I \right)A^{-1}.
\end{equation}
Thanks to \eqref{eq:flk_prec_mat}, computing the action of $\preccf_{\text{FLK}}^\top \nysmat \preccf_{\text{FLK}}$ avoids a multiplication by $\Kmm$, which is unavoidable if we compute the action of $\nysmat$ and solve with $\precond_{\text{FLK}}$ separately, as in the left variant. This computational exploit does not work for \algname{} due to the absence of off diagonal blocks in the preconditioner. Finally, we note that the \krill{} preconditioner $\precond_{\text{KRL}}$ \eqref{eq:krill_prec} is built upon the same logic as $\precond_{\text{FLK}}$, using a sparse random sign embedding~\citep{random_embeddings} rather than a simple row sampling. However, due to the more refined sampling technique, the \krill{} preconditioner does not have the same algorithmic benefits as \falkon{}, such as the double in-place Cholesky technique~\eqref{eq:flk_prec_decomposition} or the cheaper action of the preconditioned matrix~\eqref{eq:flk_prec_mat}.

\section{Datasets, algorithms and experiments details}
\label{apx:experiment_details}

\subsection{Datasets}
\label{apx:datasets_details}
For each dataset, we report the dimensions, that is, the total number of data $N$ and the number of features $d$, the preprocessing details and the error metric employed to test the model. Except for \emph{MillionSongs}, we always report mean and standard deviation using 5-fold cross-validation, therefore the actual number of training data is $n = \frac{4}{5}N$, and the amount of test data is $n_\mathrm{test}=\frac{1}{5}N$.
\paragraph{Taxi} 
A regression dataset with dimensions $N=1.1\times 10^9, d=9$. Data can be downloaded by following the instructions at \url{https://github.com/toddwschneider/nyc-taxi-data}, and is then preprocessed following \citet{peng2017asynchronous,falkon_paper2}. We report mean and standard deviation of the RMSE.
\paragraph{Cifar10-5M}
A balanced classification dataset with 10 classes and dimensions $n=6\times 10^6, d=1280$ obtained by extracting the features from the Cifar10-5M dataset~\citep{nakkiran2020deep} which can be downloaded from \url{https://github.com/preetum/cifar5m}, using a pretrained MobileNetv2 network~\citep{sandler2018mobilenetv2} (the same procedure used in \citet{eigenpro4}, but note that this is different from the dataset used in \citet{joker} where raw pixels are used as features). The data is then standardized to zero mean and unit standard deviation and classes are one-hot encoded. We report mean and standard deviation of the classification error.
\paragraph{Higgs}
A binary classification dataset with dimensions $N=1.1\times 10^7, d=28$. The data can be downloaded from \url{https://archive.ics.uci.edu/ml/datasets/HIGGS}; preprocessing follows \citet{falkon_paper2}. We report mean and standard deviation of one minus the area under the ROC curve (AUC).
\paragraph{CometMC}
A heavily unbalanced binary classification dataset with dimensions $N=7.6\times 10^6, d=4$ coming from the high-energy physics community~\citep{comet_collaboration_website}. We used the same dataset as \citet{askotch}, downloaded from OpenML (dataset ID: 23397). The data was standardized to zero mean and unit standard deviation. To account for class imbalance, and following the original Kaggle competition where the dataset comes from we report the mean and standard deviation of the $1-\mathrm{AUC}$.
\paragraph{Airlines}
A regression dataset with dimensions $N=5.9\times 10^6,d=8$. We used the same preprocessing steps as \citet{falkon_paper2} and report the mean and standard deviation of the MSE.
\paragraph{MillionSongs}
A regression dataset~\citep{Bertin-Mahieux2011} with dimensions $N = 5.1\times 10^5,d = 90$. Data can be downloaded from \url{https://archive.ics.uci.edu/ml/datasets/YearPredictionMSD}. Both features and labels were normalized to have zero mean and unit standard deviation. The error reported is the \emph{relative error} 
\[\sqrt{\frac{1}{n_\mathrm{test}}\sum_{i=1}^{n_\mathrm{test}}\left(\frac{Y_{\mathrm{true}}^{(i)} - Y_{\mathrm{pred}}^{(i)}}{Y_{\mathrm{true}}^{(i)}}\right)^2}\]
over a predefined test set of $n_\mathrm{test} = 51630$ points. We do not perform cross-validation to avoid misleading high test performances due to \say{the producer effect}.
\paragraph{Benzene}
A regression dataset from computational chemistry with dimensions $N=6.3\times 10^5, d=66$. The data was originally published in \citet{chmiela2017machine} and is available for download at \url{https://sgdml.org/}. Preprocessing follows \citet{askotch} as well as the use of the Mat{\'e}rn kernel. The error reported is on the MSE of the potential energy, calculated across 5 folds.
\paragraph{SVHN}
A balanced multiclass classification dataset with 10 classes and dimensions $N=99289, d=1024$. The original SVHN images~\citep{netzer2011reading} are converted to black-and-white and pixels are rescaled to the 0--1 range. Classes are one-hot encoded. We report the mean and standard deviation of the classification error.
\paragraph{FashionMNIST}
A balanced multiclass classification dataset with 10 classes and dimensions $N=70000, d=784$, originally published in \citet{xiao2017}. We use raw pixels as features simply rescaled to the 0--1 range. We report the mean and standard deviation of the classification error.
\paragraph{MiniBooNE}
A mildly unbalanced binary classification dataset with  dimensions $N=130082,d=50$ coming from high-energy physics. It can be downloaded from \url{https://archive.ics.uci.edu/dataset/199/miniboone+particle+identification#}. We preprocessed the features to have zero mean and unit standard deviation and report the mean and standard deviation of the $1-\mathrm{AUC}$.

\subsection{Algorithms}
Throughout the paper we highlight the performance of \algname{} by comparing it to several state of the art kernel-based learning algorithms. We report the parameters used for each algorithm on each dataset, relatively to the results shown in \Cref{tbl:large-scale} and \Cref{tbl:small-scale}, in \Cref{tab:parameters_details}, with the meaning of the parameters $m$ and $m_0$ detailed in \Cref{tab:algorithm_comparison}. The experiments on the model scaling on HIGGS in \Cref{tab:higgs-scale} have all been run with the same parameters, changing only $m$. We set the parameters of the competitor algorithms to the default values suggested by the authors. For \joker{}, we trained both the KRR and SVM models on all datasets. The table indicates which model achieved the best test error; this is the model whose results are reported in the main paper. For \eigenpro{4}, we report both the rank of the Nystr\"om approximation and the number of approximated eigenvalues (see Appendix~\ref{apx:further_comparison}) in the format \emph{rank} / $m_0$. For \algname{}, we also report whether we store one block in memory at a time (\emph{recompute blocks}) or all blocks simultaneously (\emph{store blocks}). In our experiments we use the Gaussian, Laplacian, and Mat{\'e}rn-5/2 kernels
\begin{equation*}
\begin{aligned}
    &k_{\mathrm{Gauss}}(x, x') = \exp\left(-\frac{\|x - x'\|_2^2}{2\sigma^2}\right), \qquad k_{\mathrm{Lap}}(x, x') = \exp\left(-\frac{\|x - x'\|_1}{\sigma}\right),\\
    &k_{\mathrm{Mat-5/2}}(x, x') = \left(1 + \frac{\sqrt{5}\|x - x'\|_2}{\sigma} + \frac{5\|x - x'\|_2^2}{3\sigma^2}\right) \exp\left(-\frac{\sqrt{5}\|x - x'\|_2}{\sigma}\right),
\end{aligned}
\end{equation*}
where $\sigma$ denotes the kernel bandwidth. We note that \eigenpro{4} implements a non-standard version of the Laplacian kernel, replacing the $L^1$-norm $\| \cdot \|_1$ in the exponent with the standard Euclidean norm $\| \cdot \|_2$. The general Mat{\'e}rn-5/2 kernel also includes a signal-variance parameter, which we fix to $1$, following \citet{askotch}. For all competing methods, we used the default algorithmic parameters recommended in the respective papers. The kernel, the bandwidth $\sigma$ and the regularization parameter $\lambda$ were selected by grid search. We considered Gaussian, Laplacian, and Mat{\'e}rn-5/2 kernels; since the Mat{\'e}rn kernel is not implemented in the \joker{} and \eigenpro{4} libraries, the search for these two methods was restricted to the Gaussian and Laplacian kernels. When a dataset had already been used in one of the original papers, we searched over a small grid centered on the hyperparameters reported there. Otherwise, we centered the grid for $\sigma$ on the median heuristic~\citep{median_trick}, that is, the median of the pairwise distances between data points, estimated on a random subsample of points, and searched $\lambda$ over the grid $\{10^{-9}, 10^{-8}, 10^{-7}, 10^{-6}\}$. For \joker{}, we searched over $\lambda = 2^{-i}/n$ with $i \in \{0, \dots, 7\}$, following the grid suggested by their paper. Finally, \eigenpro{4} solves the problem in the interpolation regime, so $\lambda = 0$ by default for this method.

\subsection{Experiments}\label{apx:machine_details}
The experiments of Section~\ref{sec:experiments} were run in single precision (FP32) on one node of the Leonardo cluster hosted by CINECA, requesting $320$~GiB of RAM, $8$ Intel Ice Lake $2.60$~GHz CPU cores and a single custom
NVIDIA Ampere A100~GPU with $64$~GiB of VRAM.\footnote{See \url{docs.hpc.cineca.it/hpc/leonardo.html} for the detailed hardware specifications.}
We implemented all scripts in Python~3.11.7, and used the following libraries: NumPy~2.4.4~\citep{numpy}; PyTorch~2.7.0~\citep{pytorch} with CUDA~12.6; KeOps~2.3~\citep{keops}; Matplotlib~3.11.1~\citep{matplotlib}. We implemented the code for \algname{} inside the \falkon{} library, and ran the competitors' algorithms using their official released codes.\footnote{\falkon{}: \url{https://github.com/FalkonML/falkon}; \eigenpro{4}: \url{https://github.com/EigenPro/EigenPro/tree/main}; \joker{}: \url{https://github.com/Apple-Zhang/Joker-paper}; \askotch{}: \url{https://github.com/pratikrathore8/fast_krr}.} Except for \emph{MillionSongs}, we ran $5$-fold cross-validation for all datasets. Table~\ref{tab:results_extended} reports the training and testing time, and the test error with standard deviations, complementing the results presented in Table~\ref{tbl:large-scale} and Table~\ref{tbl:small-scale}. Similarly Table~\ref{tab:higgs-scale-extended} complements the results of Table~\ref{tab:higgs-scale}. The smaller experiments of Appendix~\ref{apx:block_prec} were run in double precision (FP64) on an ASUS ROG Strix Scar 16 laptop, with $32$ GiB of RAM, an Intel Core Ultra 9 275HX CPU and an NVIDIA GeForce RTX 5090 Laptop GPU.
\begingroup
\setlength{\tabcolsep}{6pt}
\renewcommand{\arraystretch}{1.0}
\setlength{\LTcapwidth}{\textwidth}
\scriptsize

\begin{longtable}{@{}llccccccp{2.5cm}@{}}
\caption{Parameters used for each dataset and algorithm. Here $\sigma$ denotes
the kernel bandwidth and $\lambda$ the regularization parameters. For each algorithm, the meaning of the parameters $m$ and $m_0$ is explained in Table~\ref{tab:algorithm_comparison}.}
\label{tab:parameters_details}\\
\toprule
dataset & algorithm & kernel & $\sigma$ & $\lambda$ & $m$ & $m_0$ & epochs & other \\
\midrule
\endhead
\midrule
\multicolumn{8}{r}{\itshape continued on next page}\\
\endfoot
\bottomrule
\endlastfoot
\multirow{5}{*}{Higgs}
  & \falkon      & Gaussian & 3.5 & \num{3e-8} & \num{100000} & \na & 20  & \\*
  & \joker       & Laplacian & median (\num{10.7}) & $\frac{1}{n}$ & \num{200000} & 512 & 2  & SVM \\*
  & \askotch     & Gaussian & 3.5 & \num{3e-8} & Full & \num{100} & 20  & \\*
  & \eigenpro{4} & Laplacian$^\dag$ & 15 & \num{0} & \num{500000} & \num{10000} / \num{1000} & 10  & \\*
  & \algname     & Gaussian & 3.5 & \num{3e-8} & \num{400000} & \num{100000} & \num{20}  & recompute blocks\\
\midrule
\multirow{5}{*}{COMET MC}
  & \falkon      & Gaussian & 0.25 & \num{1e-8} & \num{50000} & \na & 20  & \\*
  & \joker       & Gaussian & 0.25 & $\frac{1}{128n}$ & \num{200000} & \num{512} & 10  & SVM \\*
  & \askotch     & Gaussian & 0.25 & \num{1e-8} & Full & \num{100} & \num{20}  & \\*
  & \eigenpro{4} & Laplacian$^\dag$ & 0.1 & \num{0} & \num{400000} & \num{1000} / \num{100} & 10  & \\*
  & \algname     & Gaussian & 0.25 & \num{1e-8} & \num{400000} & \num{50000} & \num{20}  & recompute blocks \\
\midrule
\multirow{5}{*}{CIFAR10-5M}
  & \falkon      & Gaussian & 27 & \num{1e-9} & \num{100000} & \na & 10  & \\*
  & \joker       & Gaussian & 27 & $\frac{1}{128n}$ & \num{400000} & \num{512} & 5  & SVM \\*
  & \askotch     & \multicolumn{7}{c}{Exceeds time limits}  \\*
  & \eigenpro{4} & Gaussian & 27 & \num{0} & \num{400000} & \num{10000} / \num{1000} & 5  & \\*
  & \algname     & Gaussian & 27 & \num{1e-9} & \num{400000} & \num{100000} & 10  & recompute blocks \\
\midrule
\multirow{5}{*}{Taxi}
  & \falkon      & Gaussian & 1 & \num{1e-8} & \num{100000} & \na & 10  & $\mathrm{PC}_\varepsilon = 10^{-7}$\\*
  & \joker       & Gaussian & 1 & $\frac{1}{128n}$ & \num{100000} & \num{512} & 1  & KRR \\*
  & \askotch     & \multicolumn{7}{c}{Exceeds time limits} \\*
  & \eigenpro{4} & \multicolumn{7}{c}{Exceeds time limits} \\*
  & \algname     & Gaussian & 1 & \num{1e-8} & \num{200000} & \num{100000} & 15  & $\mathrm{PC}_\varepsilon = 10^{-7}$,\newline recompute blocks \\
\midrule
\multirow{5}{*}{MillionSongs}
  & \falkon      & Gaussian & 7 & \num{2e-6} & \num{50000} & \na & 20 &  \\*
  & \joker       & Gaussian & median ($10.8$) & $\frac{1}{n}$ & Full & 2048 & 20 & KRR \\*
  & \askotch     & Gaussian & 7 & \num{2e-6} & Full & 100 & 20 &  \\*
  & \eigenpro{4} & Laplacian$^\dag$ & 6 & \num{0} & \num{400000} & \num{1000} / \num{100} & 20 &  \\*
  & \algname     & Gaussian & 7 & \num{2e-6} & \num{200000} & \num{50000} & 20 & store blocks  \\
\midrule
\multirow{5}{*}{Airlines}
  & \falkon      & Laplacian & median (\num{3.48}) & \num{1e-7} & \num{100000} & \na & 20 &  \\*
  & \joker       & Laplacian & median (\num{3.48}) & $\frac{1}{32n}$ & \num{200000} & 2048 & 10 & KRR \\*
  & \askotch     & Laplacian & median (\num{3.48}) & \num{1e-8} & Full & \num{100} & 20 &  \\*
  & \eigenpro{4} & Laplacian$^\dag$ & 5 & \num{0} & \num{200000} & \num{10000} / \num{1000} & 10 &  \\*
  & \algname     & Laplacian & median (\num{3.48}) & \num{1e-8} & \num{200000} & \num{100000} & 20 & recompute blocks \\
\midrule
\multirow{5}{*}{Benzene}
  & \falkon      & Mat{\'e}rn 5/2 & 8.124 & \num{1e-9} & \num{50000} & \na & 20 &  \\*
  & \joker       & Gaussian & 5 & $\frac{1}{128n}$ & \num{100000} & \num{2048} & 10 & KRR \\*
  & \askotch     & Mat{\'e}rn 5/2 & 8.124 & \num{1e-9} & Full & 100 & 20 &  \\*
  & \eigenpro{4} & Laplacian$^\dag$ & 16 & \num{0} & \num{100000} & \num{10000} / \num{1000} & 10 &  \\*
  & \algname     & Mat{\'e}rn 5/2 & 8.124 & \num{1e-9} & \num{260000} & \num{26000} & 30 & store blocks \\
\midrule
\multirow{5}{*}{MiniBooNE}
  & \falkon      & Gaussian & 1 & \num{1e-7} & \num{50000} & \na & 10 &  \\*
  & \joker       & Gaussian & 1 & $\frac{1}{4n}$ & Full & \num{2048} & 10 & SVM \\*
  & \askotch     & Gaussian & median ($2.35$) & \num{1e-9} & Full & 100 & 20 &  \\*
  & \eigenpro{4} & Laplacian$^\dag$ & 4 & \num{0} & \num{100000} & \num{10000} / \num{1000} & 20 &  \\*
  & \algname     & Gaussian & 1 & \num{1e-7} & \num{100000} & \num{50000} & 15 & store blocks \\
\midrule
\multirow{5}{*}{SVHN}
  & \falkon      & Gaussian & 3 & \num{1e-6} & \num{50000} & \na & \num{20} &  \\*
  & \joker       & Gaussian & 3 & $\frac{1}{4n}$ & \num{50000} & 2048 & 10 & SVM \\*
  & \askotch     & Gaussian & 3 & \num{1e-6} & Full & \num{100} & 10 & 10 one-vs-all models \\*
  & \eigenpro{4} & Gaussian & 4 & \num{0} & \num{50000} & \num{10000} / \num{1000} & \num{10} &  \\*
  & \algname     & Gaussian & 3 & \num{1e-6} & \num{50000} & \num{25000} & \num{20} & store blocks \\
\midrule
\multirow{5}{*}{FashionMNIST}
  & \falkon      & Gaussian & 5 & \num{1e-6} & \num{50000} & \na & \num{20} &  \\*
  & \joker       & Gaussian & 5 & $\frac{1}{8n}$ & \num{50000} & 2048 & 10 & SVM \\*
  & \askotch     & Gaussian & 5 & \num{1e-6} & Full & \num{100} & 10 & 10 one-vs-all models \\*
  & \eigenpro{4} & Gaussian & 4 & \num{0} & \num{50000} & \num{10000} / \num{1000} & \num{10} &  \\*
  & \algname     & Gaussian & 5 & \num{1e-6} & \num{50000} & \num{10000} & \num{20} & store blocks \\
\end{longtable}

\endgroup

\begingroup
\setlength{\tabcolsep}{4pt}
\renewcommand{\arraystretch}{1.0}
\setlength{\LTcapwidth}{\textwidth}
\scriptsize

\begin{longtable}{@{}llccc@{\hspace{1.5em}}llccc@{}}
\caption{Training time, test time and test error for each dataset and algorithm. For the test error we report the mean and standard deviation over $5$ data splits, except for MillionSongs. The test error metric depends on the dataset: RMSE (Taxi), MSE (Airlines, Benzene), $1-\mathrm{AUC}$ (Higgs, COMET MC, MiniBooNE), classification error (CIFAR10-5M, FashionMNIST, SVHN) and relative error (MillionSongs).}
\label{tab:results_extended}\\
\toprule
          &           & train    & test     & test  &         &           & train    & test     & test  \\
dataset   & algorithm & time (s) & time (s) & error & dataset & algorithm & time (s) & time (s) & error \\
\midrule
\endhead
\midrule
\multicolumn{10}{r}{\itshape continued on next page}\\
\endfoot
\bottomrule
\endlastfoot
\multirow{5}{*}{Taxi}
  & \falkon      & $\num{8970.89}$    & $\num{61.20}$  & $\num{307.9} \pm \num{0.1}$     & \multirow{5}{*}{MillionSongs} & \falkon      & $\num{39.69}$      & $\num{0.12}$   & $\num{0.004492}$                \\*
  & \joker       & $\num{25131.92}$   & $\num{541.45}$ & $\num{307.8} \pm \num{0.5}$     &                               & \joker       & $\num{537.77}$     & $\num{2.04}$   & $\num{0.004510}$                \\*
  & \askotch     & \multicolumn{3}{c}{Exceeds time limits}                               &                               & \askotch     & $\num{809.03}$     & $\num{1.14}$   & $\num{0.004475}$                \\*
  & \eigenpro{4} & \multicolumn{3}{c}{Exceeds time limits}                                               &                               & \eigenpro{4} & $\num{1151.67}$    & $\num{1.78}$   & $\num{0.004502}$                \\*
  & \algname     & $\num{18330.89}$   & $\num{121.54}$ & $\num{307.0} \pm \num{0.1}$     &                               & \algname     & $\num{254.47}$     & $\num{0.46}$   & $\num{0.004471}$                \\
\midrule
\multirow{5}{*}{Higgs}
  & \falkon      & $\num{413.11}$     & $\num{1.74}$   & $\num{0.1806} \pm \num{0.0004}$ & \multirow{5}{*}{Benzene}      & \falkon      & $\num{27.12}$      & $\num{0.14}$   & $\num{0.0160} \pm \num{0.0008}$ \\*
  & \joker       & $\num{2839.96}$    & $\num{10.88}$  & $\num{0.1800} \pm \num{0.0002}$ &                               & \joker       & $\num{219.78}$     & $\num{0.35}$   & $\num{0.0172} \pm \num{0.0018}$ \\*
  & \askotch     & $\num{15648.91}$   & $\num{143.83}$ & $\num{0.1763} \pm \num{0.0006}$ &                               & \askotch     & $\num{676.30}$     & $\num{1.30}$   & $\num{0.0016} \pm \num{0.0000}$ \\*
  & \eigenpro{4} & $\num{15419.25}$   & $\num{117.73}$ & $\num{0.1896} \pm \num{0.0002}$ &                               & \eigenpro{4} & $\num{428.91}$     & $\num{1.14}$   & $\num{0.0199} \pm \num{0.0003}$ \\*
  & \algname     & $\num{6268.30}$    & $\num{6.89}$   & $\num{0.1746} \pm \num{0.0003}$ &                               & \algname     & $\num{263.51}$     & $\num{0.68}$   & $\num{0.0015} \pm \num{0.0000}$ \\
\midrule
\multirow{5}{*}{Airlines}
  & \falkon      & $\num{103.06}$     & $\num{0.25}$   & $\num{0.720} \pm \num{0.004}$   & \multirow{5}{*}{MiniBooNE}    & \falkon      & $\num{17.69}$      & $\num{0.04}$   & $\num{0.0267} \pm \num{0.0013}$ \\*
  & \joker       & $\num{4019.92}$    & $\num{5.51}$   & $\num{0.718} \pm \num{0.005}$   &                               & \joker       & $\num{63.43}$      & $\num{0.18}$   & $\num{0.0283} \pm \num{0.0007}$ \\*
  & \askotch     & $\num{1788.65}$    & $\num{12.01}$  & $\num{0.717} \pm \num{0.007}$   &                               & \askotch     & $\num{292.08}$     & $\num{0.12}$   & $\num{0.0259} \pm \num{0.0013}$ \\*
  & \eigenpro{4} & $\num{6343.74}$    & $\num{26.08}$  & $\num{0.791} \pm \num{0.003}$   &                               & \eigenpro{4} & $\num{329.55}$     & $\num{0.23}$   & $\num{0.0311} \pm \num{0.0003}$ \\*
  & \algname     & $\num{2846.39}$    & $\num{0.57}$   & $\num{0.706} \pm \num{0.004}$   &                               & \algname     & $\num{40.43}$      & $\num{0.08}$   & $\num{0.0282} \pm \num{0.0013}$ \\
\midrule
\multirow{5}{*}{CIFAR10-5M}
  & \falkon      & $\num{868.86}$     & $\num{19.27}$  & $\num{12.3}\% \pm \num{0.0}\%$  & \multirow{5}{*}{FashionMNIST} & \falkon      & $\num{30.41}$      & $\num{0.09}$   & $\num{8.8}\% \pm \num{0.2}\%$   \\*
  & \joker       & $\num{5391.51}$    & $\num{67.50}$  & $\num{12.3}\% \pm \num{0.0}\%$  &                               & \joker       & $\num{32.39}$      & $\num{0.08}$   & $\num{9.6}\% \pm \num{0.1}\%$   \\*
  & \askotch     & \multicolumn{3}{c}{Exceeds time limits}                               &                               & \askotch     & $\num{4344.72}$    & $\num{4.65}$   & $\num{9.0}\% \pm \num{0.1}\%$   \\*
  & \eigenpro{4} & $\num{5375.92}$    & $\num{92.42}$  & $\num{13.1}\% \pm \num{0.1}\%$  &                               & \eigenpro{4} & $\num{76.41}$      & $\num{0.14}$   & $\num{9.2}\% \pm \num{0.2}\%$   \\*
  & \algname     & $\num{6738.94}$    & $\num{76.16}$  & $\num{12.0}\% \pm \num{0.0}\%$  &                               & \algname     & $\num{14.17}$      & $\num{0.10}$   & $\num{8.8}\% \pm \num{0.2}\%$   \\
\midrule
\multirow{5}{*}{COMET MC}
  & \falkon      & $\num{45.67}$      & $\num{0.13}$   & $\num{0.0089} \pm \num{0.0001}$ & \multirow{5}{*}{SVHN}         & \falkon      & $\num{66.08}$      & $\num{0.16}$   & $\num{19.6}\% \pm \num{0.2}\%$  \\*
  & \joker       & $\num{3330.70}$    & $\num{7.29}$   & $\num{0.0077} \pm \num{0.0001}$ &                               & \joker       & $\num{51.59}$      & $\num{0.13}$   & $\num{23.3}\% \pm \num{0.5}\%$  \\*
  & \askotch     & $\num{2026.83}$    & $\num{11.74}$  & $\num{0.0059} \pm \num{0.0000}$ &                               & \askotch     & $\num{4396.48}$    & $\num{10.12}$  & $\num{20.1}\% \pm \num{0.3}\%$  \\*
  & \eigenpro{4} & $\num{5722.56}$    & $\num{64.31}$  & $\num{0.0241} \pm \num{0.0015}$ &                               & \eigenpro{4} & $\num{98.28}$      & $\num{0.24}$   & $\num{22.2}\% \pm \num{0.2}\%$  \\*
  & \algname     & $\num{1839.78}$    & $\num{0.99}$   & $\num{0.0057} \pm \num{0.0002}$ &                               & \algname     & $\num{70.29}$      & $\num{0.15}$   & $\num{19.8}\% \pm \num{0.2}\%$  \\
\end{longtable}

\endgroup

\begin{table}[ht]
\centering
\caption{Training time, test time and AUC for each model size $m$ of \algname{} on the HIGGS dataset. For the AUC we report the mean and standard deviation over $5$ data splits.}
\begin{tabular}{lccccc}
\toprule
$m$ & $200\mathrm{k}$ & $400\mathrm{k}$ & $800\mathrm{k}$ & $1.6\mathrm{M}$ & $3.2\mathrm{M}$ \\
\midrule
train time (s) & $\num{3140.71}$ & $\num{6268.30}$ & $\num{12574.60}$ & $\num{25940.69}$ & $\num{52901.50}$ \\
test time (s) & $\num{3.46}$ & $\num{6.89}$ & $\num{13.76}$ & $\num{27.48}$ & $\num{54.92}$ \\
AUC  & $\num{0.8232} \pm \num{0.0003}$ & $\num{0.8254} \pm \num{0.0003}$ & $\num{0.8266} \pm \num{0.0003}$ & $\num{0.8272} \pm \num{0.0002}$ & $\num{0.8277} \pm \num{0.0003}$ \\
\bottomrule
\end{tabular}
\label{tab:higgs-scale-extended}
\end{table}

\section{Complete BLOKS pseudocode}
\label{apx:pseudocode_long}
\begin{table}[t!]
\centering
\begin{algorithm}[H]
\caption{\algname{}}
\label{alg:blk_algorithm_long}

\begin{algorithmic}[1]
    \Statex \textbf{Input}: data matrix $\widehat{X} = [x_1 \mid \dots \mid x_n]^\top \in \bbR^{n \times d}$, label vector $\widehat{y} \in \bbR^n$, penalty $\lambda > 0$, model size $m$, block size $m_0 \mid m$, kernel $\kernel$, number of iterations $t_{\max}$
    \Statex \textbf{Output}: Nystr\"om predictor $\cvn$ (approximated coefficient vector)
\end{algorithmic}
\vspace{0.2cm}

\begin{minipage}[t]{0.45\textwidth}
\begin{algorithmic}[1]
    \State $c = \texttt{sample}(\{1, \dots, n\}, m)$
    \State $\widetilde{X} = \widehat{X}[c]$
    \State $\Knm = \kernel(\widehat{X}, \widetilde{X})$
    \State $\Kmm = \Knm[c]$
    \State $r_0 = \Knm^\top \widehat{y}$ \label{line:matvec0}
    \State $z_0 = \precfuncname(r_0; \widetilde{X}, n, m, m_0)$
    \State $p_0 = z_0$
    \For{$t = 0$ to $t_{\max}-1$}
        \State $v_t = \Knm^\top (\Knm p_t) + \lambda n (\Kmm p_t)$ \label{line:matvect}
        \State $\alpha_t = \frac{r_t^\top z_t}{p_t^\top v_t}$
        \State $\cvn_{t+1} = \cvn_t + \alpha_t p_t$
        \If{$t = t_{\max}-1$} \label{line:check_last_iter}
            \State \textbf{break}
        \EndIf
        \State $r_{t+1} = r_t - \alpha_t v_t$
        \State $z_{t+1} = \precfuncname(r_{t+1}; \widetilde{X}, n, m, m_0)$ \label{line:new_prec_res}
        \State $\beta_t = \frac{r_{t+1}^\top z_{t+1}}{r_t^\top z_t}$
        \State $p_{t+1} = z_{t+1} + \beta_t p_t$
    \EndFor
    \State \Return $\cvn = \cvn_{t_{\max}}$
\algstore{main}
\end{algorithmic}
\end{minipage}%
\hfill
\begin{minipage}[t]{0.53\textwidth}
\begin{algorithmic}[1]
\algrestore{main}
    \Function{\precfuncname}{$r; \widetilde{X}, n, m, m_0$}
        \State $q = m / m_0$
        \For{$j = 1$ to $q$}
            \State $c_j = [(j - 1) \cdot m_0 : j \cdot m_0]$
            \State $\widetilde{X}_j = \widetilde{X}[c_j]$
            \State $\Kmjmj = \kernel(\widetilde{X}_j, \widetilde{X}_j)$
            \State {$\Kmjmj, \ dT_j, \ dA_j \gets \texttt{\precfuncname\_block}(\Kmjmj)$}
            \State $\texttt{diag}(\Kmjmj) \gets dT_j$
            \State $r^{(j)} \gets \texttt{triu}(\Kmjmj)^{-T}r^{(j)}$ \Comment{Solve with $T_j^{-T}$}
            \State $\texttt{diag}(\Kmjmj) \gets dA_j$
            \State $r^{(j)} \gets \texttt{tril}(\Kmjmj)^{-1}r^{(j)}$ \Comment{Solve with $A_j^{-T}$}
            \State $r^{(j)} \gets \texttt{tril}(\Kmjmj)^{-T}r^{(j)}$ \Comment{Solve with $A_j^{-1}$}
            \State $\texttt{diag}(\Kmjmj) \gets dT_j$
            \State $r^{(j)} \gets \texttt{triu}(\Kmjmj)^{-1}r^{(j)}$ \Comment{Solve with $T_j^{-1}$}
            \State \textbf{delete} $\Kmjmj, \ dT_j, \ dA_j$ \Comment{Free memory} \label{line:delete}
        \EndFor
        \State \Return $r \gets \frac{1}{nq} r$
    \EndFunction
\algstore{prec}
\end{algorithmic}
\end{minipage}

\vspace{0.5cm}

\begin{algorithmic}[1]
\algrestore{prec}
    \Function{\texttt{\precfuncname\_block}}{$\Kmjmj$}
        \State $\texttt{triu}(\Kmjmj) \gets \texttt{chol}(\Kmjmj)$ \Comment{$T_j$ stored in the upper triangular part of $\Kmjmj$}
        \State $dT_j = \texttt{diag}(\Kmjmj)$ \Comment{Diagonal of $T_j$}
        \State $\texttt{tril}(\Kmjmj) \gets \frac{1}{m_0} \texttt{triu}(\Kmjmj) \texttt{triu}(\Kmjmj)^\top + \lambda I$
        \State $\texttt{tril}(\Kmjmj) \gets \texttt{chol}(\texttt{tril}(\Kmjmj))^\top$ \Comment{$A_j^\top$ stored in the lower triangular part of $\Kmjmj$} \label{line:chol_2}
        \State $dA_j = \texttt{diag}(\Kmjmj)$ \Comment{Diagonal of $A_j$}
        \State \Return $\Kmjmj, \ dT_j, \ dA_j$
    \EndFunction
\end{algorithmic}

\end{algorithm}
\end{table}
We provide in Algorithm~\ref{alg:blk_algorithm_long} the detailed pseudocode for \algname{}, using standard Python indexing. We assume starting point $\cvn_0 = 0$. This is the standard choice if no prior information on the solution is available, and avoids a matrix vector multiplication to compute the first residual $r_0$, which is just the right-hand side $\Knm^\top \widehat{y}$ of \eqref{eq:NKRR_system} in this case. For notational clarity and compactness, the algorithm is written as if all kernel matrices were computed and stored explicitly. In practice, whenever a matrix-vector multiplication is performed (on line~\ref{line:matvec0} and line~\ref{line:matvect}), a function computing the product on the fly is invoked. On line~\ref{line:check_last_iter} there is an explicit instruction to check if the last iteration has been reached, and a manual break if that is the case. This is a minor optimization that avoids extra computations required only for the subsequent iteration, in particular the computation of the new preconditioned residual on line~\ref{line:new_prec_res}. Finally, the function \texttt{prec\_block} implements the double in-place Cholesky decomposition described in \Cref{sec:algorithm}. Note that the total memory requirement is that of the single block $\Kmjmj$, plus the negligible space required to store the diagonals $dT_j$ and $dA_j$. With an abuse of notation, when a matrix is completely determined by its upper/lower triangular part (that is, when it is an upper/lower triangular or symmetric matrix), and we assign it to the upper (\texttt{triu}) or lower (\texttt{tril}) portion of a matrix in memory, it means that we are overwriting the upper/lower triangular part of one with the upper/lower triangular part of the other. Furthermore, when we perform the Cholesky decomposition of a triangular matrix as in line \ref{line:chol_2}, it means that we are performing the Cholesky decomposition of the corresponding symmetric (and positive definite) matrix.

This implementation corresponds to the regime in which blocks are too large to fit in memory and must be recomputed from scratch at each iteration. If all blocks can be stored in memory simultaneously, the algorithm is the same, except that line~\ref{line:delete} is absent and the blocks are precomputed before the
optimization loop. In both regimes, the blocks are processed sequentially, as all our experiments are run on a single GPU. Since the block solves are independent, they could be distributed across multiple GPUs, so the training times reported in the main text should be seen as conservative. Even in this setting, as shown in \Cref{sec:experiments}, \algname{} achieves training times competitive with the state of the art.

\end{document}